\pdfoutput=1

\documentclass[11pt]{article}

\usepackage[]{EMNLP2022}

\usepackage{times}
\usepackage{latexsym}

\usepackage[T1]{fontenc}

\usepackage[utf8]{inputenc}

\usepackage{microtype}

\usepackage{inconsolata}

\usepackage{times}
\usepackage{latexsym}
\usepackage{tcolorbox}
\usepackage{multirow}
\usepackage{tikz}
\usepackage{capt-of}
\usepackage{graphicx}  
\usepackage{pgfplots}
\pgfplotsset{compat=1.12}
\usepackage{amsmath}
\usepackage{multicol}
\usepackage{booktabs}
\usepackage{color}
\usepackage{colortbl,array,xcolor}
\usepackage{xspace}

\usetikzlibrary{intersections}

\newcommand{\Sref}[1]{Section~\ref{#1}}
\newcommand{\Tref}[1]{Table~\ref{#1}}
\newcommand{\Fref}[1]{Figure~\ref{#1}}
\newcommand{\Eref}[1]{Eq.~\ref{#1}}
\usepackage{caption}
\usepackage{subcaption}
\usepackage{graphicx}
\usepackage{amsfonts}
\usepackage{booktabs}

\definecolor{Gray}{gray}{0.85}
\definecolor{LightCyan}{rgb}{0.88,1,1}

\newcolumntype{a}{>{\columncolor{Gray}}c}
\newcolumntype{b}{>{\columncolor{LightCyan}}c}

\usetikzlibrary{shapes.geometric}
\usepackage{xspace}

\usepackage{amsmath}

\usepackage{algorithm}
\usepackage{algpseudocode}%

\newcommand{\ours}{{\sc Attempt}}
\usepackage{amsfonts,amssymb}
\usepackage{soul}

\title{{\ours}: Parameter-Efficient Multi-task Tuning \\ via Attentional Mixtures of Soft Prompts }
 \author{
{\centering Akari Asai$^\heartsuit$~ 
Mohammadreza Salehi$^\heartsuit$~
  Matthew E. Peters$^\diamondsuit$~
  Hannaneh Hajishirzi$^{\heartsuit\diamondsuit}$ } \\
 $^\heartsuit$ University of Washington~~$^\diamondsuit$ Allen Institute for AI \\
  \texttt{\{akari, mrsalehi, hannaneh\}@cs.washington.edu}  \\
  \texttt{matthewp@allenai.org} \\
}

\begin{document}
\maketitle
\begin{abstract}
This work introduces a new multi-task, parameter-efficient language model (LM) tuning method that learns to transfer knowledge across different tasks via a mixture of soft {\it prompts}---small prefix embedding vectors pre-trained for different tasks. 
{Our method, called {\ours} ({\textsc ATTE}ntional {\textsc M}ixtures of {\sc P}rompt {\sc T}uning), obtains {\it source prompts} as encodings of large-scale source tasks into a small number of parameters and trains an attention module to interpolate the source prompts and a newly initialized target prompt for every instance in the target task. 
During training, only the target task prompt and the attention weights, which are shared between tasks in multi-task training, are updated, while the original LM and source prompts are intact. }
\ours~is highly parameter-efficient (e.g., updates  2,300 times fewer parameters than {{full}} fine-tuning), while achieving high task performance using knowledge from high-resource tasks. 
Moreover, it is modular using pre-trained soft prompts and can flexibly add or remove source prompts for effective knowledge transfer. 
Our experimental results across 21 diverse NLP datasets show that \ours~significantly outperforms prompt tuning and outperforms or matches fully fine-tuned or other parameter-efficient tuning approaches that use over ten times more parameters. Finally,  {\ours} outperforms previous work in few-shot learning settings.\footnote{Our code is available at \url{https://github.com/AkariAsai/ATTEMPT}. } 
\end{abstract}

\section{Introduction}
Fine-tuning all the parameters of large language models (LMs) given target task training data is the most common practice for optimizing task performance~\cite{devlin-etal-2019-bert,raffel2019exploring}. 
A recent line of research introduces parameter-efficient tuning methods~\cite{pmlr-v97-houlsby19a,li-liang-2021-prefix,zaken2021bitfit} that only update a small number of LM parameters; however, increasing efficiency often decreases the task performance~\cite{he2021towards}.
Moreover, these models are trained only using the task training data and  do not benefit from large collection of other NLP tasks~\cite{liu-etal-2019-multi}. We posit that parameter-efficient tuning methods  can leverage  rich knowledge of high-resource tasks to improve both training efficiency and task performance.

\begin{figure}[t!]
\includegraphics[width=8cm]{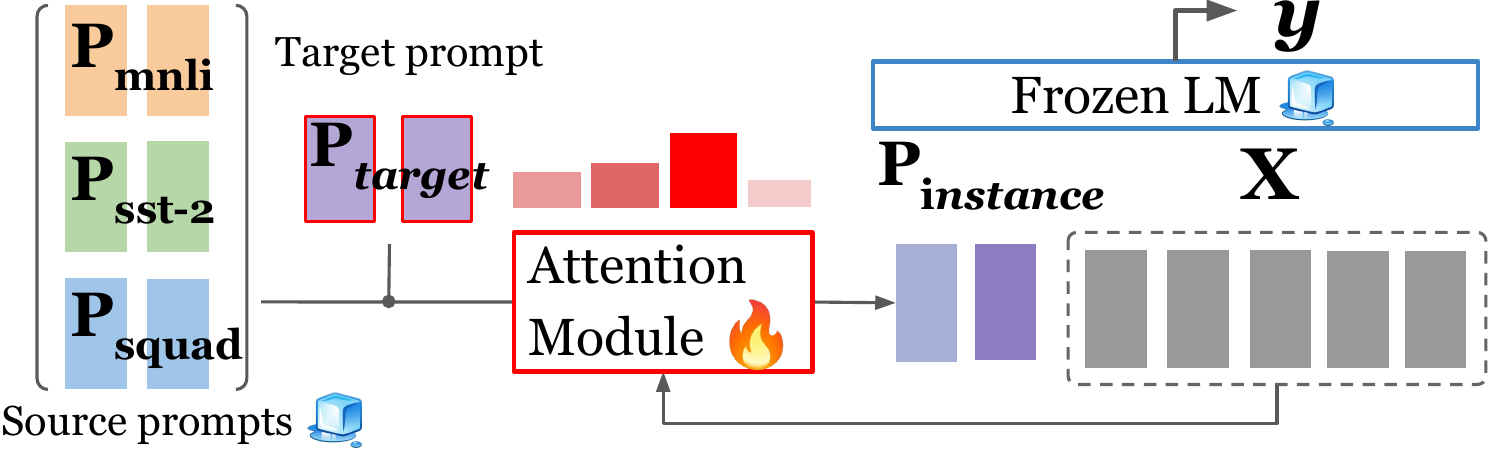}\caption{{\ours} combines multiple soft prompts trained on large-scale datasets ({\it source prompts}) to generate instance-wise prompts for a target task. At target task training, the LM and source prompts are intact. 
} \label{attempt_teaser}
\end{figure}

This work introduces a new parameter-efficient, modular multi-task tuning method called {\ours} ({\sc ATTE}ntional {\sc M}ixtures of {\sc P}rompt {\sc T}uning, previewed in \Fref{attempt_teaser}). \ours~efficiently integrates knowledge from multiple tasks via a mixture of trainable {soft prompts} preprended to the input, keeping the original LM completely frozen.
It first {pre-trains transferable soft embeddings~\cite{lester-etal-2021-power}, called {\it source prompts}, on large-scale source tasks, which are likely to contain knowledge beneficial to other tasks. 
}
Then, \ours~initializes a new {\it target prompt} for a given target task and learns an attention-weighted combination of source prompts and the target prompt.
The attention module is a light-weight network that can be shared and trained simultaneously across tasks. 

{
{\ours} offers three key advantages over previous multi-task fine-tuning or parameter-efficient tuning methods: first, it is \textbf{highly parameter-efficient }and achieves competitive performance despite updating only 0.4\% of the  parameters in full fine-tuning. Second, it enables \textbf{modular multi-task learning} using pre-trained soft prompts, where knowledge from different tasks can be flexibly combined, reused, or removed, and new tasks can be added to the lists of source or target tasks. 
{Unlike prior work that relies on precomputed priors on which tasks are related, \ours~learns to focus on useful tasks from many source tasks. }
Moreover, at inference, a single LM with multiple pre-loaded soft prompts can perform multiple tasks.
Lastly, it \textbf{improves interpretability} on underlying task similarities in multi-task learning by generating attention distributions. }

We conduct experiments on 21 datasets across diverse tasks, domains and output formats.
{\ours} significantly outperforms previous prompt tuning-based approaches by a large margin. Moreover, it matches state-of-the-art parameter-efficient transfer approaches or fully fine-tuned models that train orders of magnitude more parameters, especially on smaller datasets. \ours~is also effective on few-shot domain adaptations (i.e., 4-32 shots). 

Our analysis further shows that \ours~is particularly parameter-efficient and competitive with larger backbone LMs, where other parameter-efficient transfer approaches show rapid increases of the trainable parameters. 
Our ablation studies suggest that learned attentions, multi-task learning and modular transfer from multiple tasks largely contribute to the performance improvements. 
The attention distributions show the underlying similarities among seemingly different tasks (e.g., entailment and paraphrase detection), indicating signal for effective knowledge transfer across tasks. 
\section{Background and Problem Setup}
We first enlist common  paradigms in NLP for learning a  target task, which differ in terms of  available data and  resources. We then describe our problem setup with respect to these paradigms. 

\paragraph{Fine-tuning.}~~The most common practice in learning a new target task $T_{target}$ is to fine-tune all parameters of a pre-trained LM on the target task training data $\{(\boldsymbol{x},\boldsymbol{y})\}$~(e.g.,~\citealt{devlin-etal-2019-bert}). 
Formally, given pre-trained LM parameters $\theta$, fine-tuning results in a specialized model  $\theta_{target}$ by optimizing:  
$\underset{\theta_{target}}{\max}~ p_{\theta_{target}}(\boldsymbol{y} \mid \boldsymbol{x})$.

\paragraph{Parameter-efficient tuning.}
To decrease training costs, parameter-efficient tuning updates a small number of parameters for the target task $\phi_{target}$: $\underset{\phi_{target}}{\max}~p_{\theta,\phi_{target}}(\boldsymbol{y} \mid \boldsymbol{x})$, where the number of  $\phi_{target}$ is much smaller than $\theta_{target}$.
Adapter~\cite{pmlr-v97-houlsby19a} and its variants~\cite{mahabadi2021compacter,ruckle-etal-2021-adapterdrop} insert trainable layers in the LMs for each task, and BitFit~\cite{zaken2021bitfit} directly updates LM biases only. 
Highly efficient prefix-tuning~\cite{li-liang-2021-prefix} and prompt tuning~\cite{lester-etal-2021-power} keep the original LM frozen and only update {\it soft prompts} prepended to the input.   
In-context learning~\cite{brown2020language} uses massive-scale LMs to learn new tasks from  demonstrations ({\it hard prompts}) without any parameter update of $\theta$, but often perform worse than the aforementioned methods with parameter updates~\cite{liu2022few}. 
SPoT~\cite{vu2021spot} demonstrates that transferring prompts to another task enhances the performance at the cost of massive search. 
Given the rapidly increasing size of pre-trained LMs~\cite{chowdhery2022palm,brown2020language}, efficiently tuning to a new target task is desirable, but it often incurs a performance cost compared to the fine-tuning methods or shows sensitivity toward initialization~\cite{li-liang-2021-prefix,lester-etal-2021-power}.  

\begin{figure*}[t!]
\includegraphics[width=16cm]{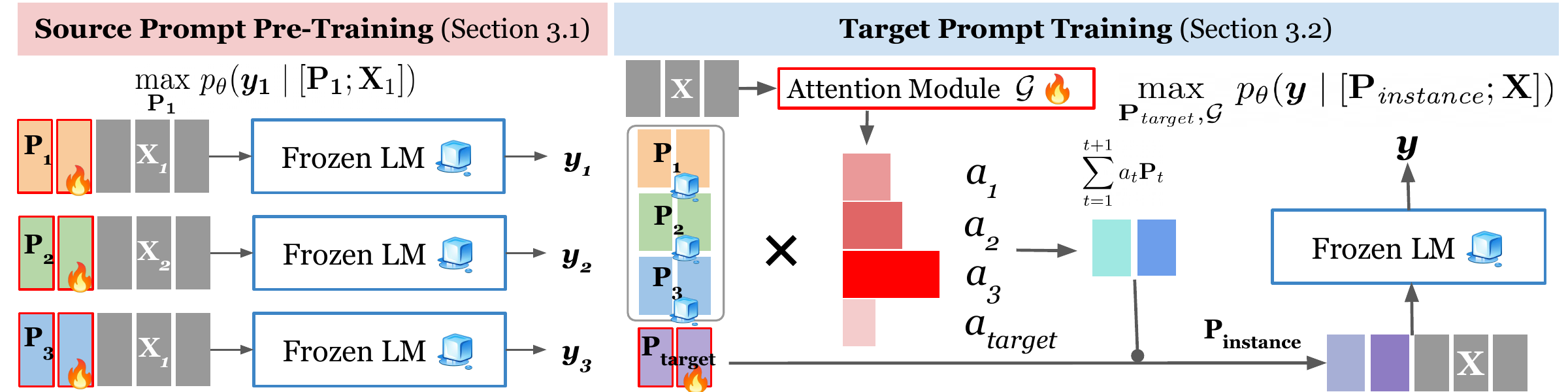}\caption{ Overview of \ours. The parts framed in red are updated during training while other parts are intact. }\label{fig:overview}
\end{figure*}

\paragraph{Multi-task transfer learning.} 
Transfer learning methods attempt to learn a new target task given a collection of source tasks by updating the parameters of an LM, which has been proven effective in NLP~\cite{khashabi-etal-2020-unifiedqa,raffel2019exploring}. 
Common approaches train on 
many different tasks~\cite{liu-etal-2019-multi,aribandi2021ext5}
or transfer a model fine-tuned on source tasks to another target task~\cite{vu-etal-2020-exploring,talmor-berant-2019-multiqa}. 
{
Several recent work introduce zero-shot or few-shot transfer of massive multi-task pretrained models~\cite{sanh2022multitask,min2021metaicl,wang2022benchmarking,mishra-etal-2022-cross,wei2022finetuned}{, without any parameter updates.}
However, those massive multi-task training approaches lack the flexibility of adding or removing source tasks ({\it modularity}) even when some tasks cause negative interference between competing tasks~\cite{zhang2020survey, aghajanyan-etal-2021-muppet}.}

\paragraph{Our problem setup.}
We combine parameter-efficient tuning and multi-task learning. 
Given a collection of source tasks $T_1, \ldots T_{t}$, our goal is to learn a new task $T_{target}$ by efficiently updating parameters $\phi_{target}$ 
given the target task data $\{(\boldsymbol{x},\boldsymbol{y})\}$, transferring knowledge from the source tasks.
Importantly, we do not know a priori which tasks provide useful inductive bias in the new target task~\cite{ponti2022combining}: seemingly different  tasks can benefit from each other.

\section{Method}
\label{sec:mop}
{\sc \ours} (depicted in \Fref{fig:overview}) leverages highly parameter-efficient prompt tuning~\cite{lester-etal-2021-power} to obtain {\it source prompts} that encode knowledge from source tasks into a small number of parameters. 
It tunes instance-level prompts by integrating  the {\it source prompts} and a {\it target prompt} newly initialized for a target task through an attention mechanism for every target task instance.

\ours~pre-trains a set of source prompts $\mathbf{P}_1, \ldots, \mathbf{P}_t$ for  source tasks (\Sref{sec:source_target}; left side of \Fref{fig:overview}) and initializes a target prompt $\mathbf{P}_{\it target}$ for the target task. 
It then computes attentions between embedded input $\mathbf{X}$ and the soft prompts for each instance  $(\boldsymbol{x},\boldsymbol{y})$ using an attention module  $\mathcal{G}$  (\Sref{sec:attention_gen}). 
Subsequently, \ours~produces instance-wise prompt $\mathbf{P}_{\it instance}$ by interpolating the source prompts and the target-task prompt given the computed attentions (\Sref{sec:interpolate}). 
$\mathbf{P}_{\it instance}$ is then prepended to the input to form the final input to a frozen LM $\theta$.  

During training, \ours~only updates the weights of $\mathbf{P}_{\it target}$ and $\mathcal{G}$ by maximizing the probability of generating $\boldsymbol{y}$ given $\mathbf{P}_{\it instance}$ and $\boldsymbol{x}$. 
Importantly, it uses the unique characteristic of prompt or prefix tuning, where task-specific parameters $\phi_{\it target}$ for different tasks can be trained in the same minibatch~\cite{lester-etal-2021-power,li-liang-2021-prefix}. 
Hence, it can train a shared attention $\mathcal{G}$ and multiple target task prompts simultaneously for further parameter and inference efficiency~(\Sref{sec:multi_task}).
{Finally, we discuss parameter efficiency of \ours~in \Sref{sec:parameter_efficiency_method}}.

\subsection{{Source Prompt Pre-training}}
\label{sec:source_target}

We first obtain source prompts 
$[\mathbf{P}_1, \ldots, \mathbf{P}_t]$  for $t$ high-resource datasets, such as Multi-NLI~\cite{williams-etal-2018-broad}, SQuAD~\cite{rajpurkar-etal-2016-squad} through prompt tuning~\cite{lester-etal-2021-power}. 
Each source prompt is only trained once for a source task and can be transferred to different target tasks. 
Formally, for an input sequence $\mathbf{X}$, 
a soft prompt is represented as $\mathbf{P} = [\mathbf{p}_1, \ldots, \mathbf{p}_m] \in \mathbb{R}^{m \times d}$, where $m$ is the prompt length, and $d$ is the LM dimension.
Input embeddings prepended by the prompt $[\mathbf{P};\mathbf{X}]$ are fed into the frozen LM $\theta$. During training, only prompt embeddings are updated by maximizing the likelihood of generating the target sequence $\boldsymbol{y}$, as follows:
\begin{equation}\label{eq:prompt}
    \underset{\mathbf{P}}{\max}~p_{\theta}(\boldsymbol{y} \mid [\mathbf{P};\mathbf{X}]).
\end{equation}

\subsection{Target Prompt Training}
\label{sec:attention}
After initializing a soft prompt for a new target task $\mathbf{P}_{\it target} (=\mathbf{P}_{t+1})$, we learn instance-wise soft prompts $\mathbf{P}_\textit{instance}$  for each instance in the target task by interpolating the source prompts and the target task prompt given attention scores generated by $\mathcal{G}$.
Similar to \Eref{eq:prompt}, we concatenate the produced instance-wise prompt to the input and train \ours~by maximizing the likelihood:
\begin{equation}\label{eq:training}
     \underset{\mathbf{P}_{\it target}, \mathcal{G}}{\max} ~p_\theta(\boldsymbol{y} \mid [\mathbf{P}_\textit{instance};\mathbf{X}]).
\end{equation} 
During training, the new task prompt $\mathbf{P}_{\it target}$ and $\mathcal{G}$ are updated via $\mathbf{P}_\textit{instance}$, while source prompts and the original LM $\theta$ are untouched to preserve the knowledge learned from prior tasks or pretraining.

\subsubsection{Input-prompt Attentions}
\label{sec:attention_gen}
{\ours} controls the influence of the set of source prompts on the instance-wise prompt by calculating input-prompt attentions. 
Specifically, an attention module $\mathcal{G}$ generates the attention weights $a_1, \ldots, a_{t+1}$ from input $\mathbf{X}$ to the prompts including both source prompts and the new target prompt.

Since the input $\mathbf{X} \in \mathbb{R}^{l \times d}$ and a soft prompt $\mathbf{P}_j \in \mathbb{R}^{m \times d}$ have different sequence lengths, 
we first perform the max-pool operation for each dimension on $\mathbf{X}$ and each source prompt embedding and obtain $\hat{\mathbf{X}}\in \mathbb{R}^{d}$ and $\hat{\mathbf{P}_j} \in \mathbb{R}^{d}$.\footnote{{This does not add any new parameters, and empirically performs slightly better than other pooling approaches. }}  
We then feed $\hat{\mathbf{X}}$ to a sub-network $\mathcal{G}$ to project it into the prompt spaces.
{
\citet{khashabi2021prompt} suggest that soft prompts may not correspond to any meaningful tokens in the input embedding spaces, so simply computing similarities between $\hat{\mathbf{X}}$ and $\hat{\mathbf{P}_j}$ may not give reliable scores. 
}
For efficiency, $\mathcal{G}$ consists of down and up projection layers, as follows:
\begin{align*}
    \mathbf{H}&_{down}  &=&  \mathbf{W}_{down}^\top(\hat{\mathbf{X}}) \\
    \mathbf{H}&_{up}  &=& \mathbf{W}_{up}^\top({\rm NonLinear(}\mathbf{H}_{down}) ) \\
    \mathbf{H}&_{out} &=& {\rm LayerNorm}(\mathbf{H}_{up}),
\end{align*}
where $\mathbf{W}_{down} \in \mathbb{R}^{d \times r} (r < d)$ and $\mathbf{W}_{up}  \in \mathbb{R}^{r \times d}$ are projection parameters to be updated during training. 
We use SiLU~\cite{elfwing2017sigmoid} for the non-linear layer and apply Layer Norm~\cite{ba2016layer} on $\mathbf{H}_{up}$, observing that  without layer norm, $\mathbf{H}_{up}$ often grows quickly and gradients explode.

Finally, we compute the attentions by calculating the product between $\hat{\mathbf{P}}_j$ and $\mathbf{H}_{out}$, and apply softmax over the prompts, as follows:
\begin{equation}\label{eq:soft_max}
    a_{j} = \frac{e^{\hat{\mathbf{P}_j} \mathbf{H}_{out}} / T }{\sum_{k=1}^{t+1} e^{\hat{\mathbf{P}_k}  \mathbf{H}_{out}} / T }, 
\end{equation}
where $T$ is a softmax temperature~\cite{radford2021learning} and scale the logits in \Eref{eq:soft_max} to avoid making the attention module over-confident.

\subsubsection{Prompt Interpolation}
\label{sec:interpolate}
The final soft prompt for the instance $\mathbf{X}$ is calculated as the weighted sum of the prompts given the attention generated by \Eref{eq:soft_max}:
\begin{equation}
\label{eq:final_soft}
    {\mathbf{P}_\textit{instance}(\mathbf{X}) = \mathbf{P}_{\it target} + \sum_{j=1}^{t+1} a_j \mathbf{P}_j}.
\end{equation}
The second term on the right differs for different instances of the same task, while the $\mathbf{P}_{\it target}$ term is task-specific. 
The attentions act as a gate to control the influences from different prompts and enable a flexible composition of knowledge from multiple tasks. 
As shown in \Eref{eq:final_soft}, the selection of  $1+a_{t+1}$ weights for the target-task-specific prompt $\mathbf{P}_{\it target}(=\mathbf{P}_{t+1})$ enables {\ours} to {\it down-play} the role of  source prompts if the knowledge from none of the sources tasks is useful for the instance $\mathbf{X}$, while always keeping the influence of $\mathbf{P}_{\it target}$ so that it will be properly updated during training.

\subsection{Multi-task Training and Inference}
\label{sec:multi_task}
{
\paragraph{Training.}
\ours~can jointly train the attention module $\mathcal{G}$ and multiple target task prompts.
Here, we explain our approach on multi-task learning over a group of target tasks by sharing $\mathcal{G}$. }

It first concatenates the training datasets, while keeping each task ID information. 
During training, we retrieve the target-task prompt corresponding to the instance given the task ID, calculate attentions over the set of the prompts and produce instance-wise prompt as described in \Sref{sec:attention}.  
The loss for each target task prompt only backpropagates when the prompt is used, while the weights of the attention module is updated at each iteration.

This way, target tasks are loosely connected and together contribute to an improved and task-agnostic attention module, which is particularly effective when the target task training data is small. {Moreover, this reduces the number of parameters to be updated per task and improves the efficiency of inference time. 
}

\paragraph{Inference.}
At inference time, we load source prompts, all of the target task prompts and the shared $\mathcal{G}$ just once. 
For each instance, {\ours} retrieves the target task prompt and produces $\mathbf{P}_{instance}$ as in ~\Eref{eq:final_soft}, and then concatenates $\mathbf{P}_{instance}$ to the input embedding. The inference process after producing instance prompt is exactly the same as in prompt tuning.

{\ours~enables loading multiple target task prompts and performing multiple target tasks simultaneously, significantly reducing the inference time model loading overhead. 
Existing approaches such as full fine-tuning or Adapter requires model loading for different target tasks, making its multi-task inference pipeline complicated. }

\subsection{Parameter Efficiency of \ours}
\label{sec:parameter_efficiency_method}
For each task, we will introduce a new trainable soft prompt $m \times d$, where $m$ is the length of the prompts and $d$ is the LM's dimension.
An attention module consists of two projection matrices and a layer norm, resulting in $d\times r + r \times d  + 2d = 2rd + 2d$ parameters, where $r$ is the projection dimension.  
As this can be shared across $N$ target tasks, the additional parameters per task will be: $d \times m + \frac{2rd + 2d}{N} = d(m + 2(r+1)/N)$. 
{
A unique characteristic of {\ours} or prompt tuning is their independence from the number of the LM layers; 
With Adapter or fine-tuning, the number of the parameters quickly increases as the backbone LMs get larger. \ours, in contrast, ~updates only the soft prompts and do not modify the LM higher layers, resulting in  moderate parameter increases compared to other approaches.

When we use T5-XL as a backbone LM, Adapter and BitFit updates about 6 million and 2 million parameters respectively, while {\ours} only updates only 172k parameters per task. 
(\Fref{fig:params}).
}

\section{Experiments}

\subsection{Source and Target Tasks}
\label{sec:data_task}
We use 6 large-scale datasets as {\it source tasks}, and evaluate on {21} diverse {\it target tasks} including entailment, paraphrase detection, sentiment analysis, question answering (QA), commonsense reasoning. 
Datasets details are in Appendix \Sref{sec:dataset_appendix}. 

\paragraph{Source tasks.}
We use the following datasets with more than 100k annotations in total from GLUE, SuperGLUE and MRQA for source prompts:  MNLI~\cite{williams-etal-2018-broad}, QNLI~\cite{demszky2018transforming}, QQP~\cite{wang-etal-2018-glue}, SST-2~\cite{socher-etal-2013-recursive}, SQuAD~\cite{rajpurkar-etal-2016-squad}, and ReCoRD~\cite{zhang2018record}.

\paragraph{GLUE and SuperGLUE.}
We use 8 GLUE tasks~\cite{wang-etal-2018-glue} and 5 SuperGLUE~\cite{wang2019superglue} tasks as target datasets to test the model's natural language understanding abilities:  
BoolQ~\cite{clark-etal-2019-boolq}, CB~\cite{de2019commitmentbank}, MultiRC~\cite{khashabi-etal-2018-looking}, WiC~\cite{pilehvar-camacho-collados-2019-wic}, WSC~\cite{10.5555/3031843.3031909}, RTE~\cite{giampiccolo-etal-2007-third}, CoLA~\cite{warstadt2019neural}, STS-B~\cite{cer-etal-2017-semeval}, MRPC~\cite{dolan-brockett-2005-automatically}, MNLI, QQP, QNLI and SST-2. 
{Four of the GLUE datasets used as source tasks (MNLI, QQP, SST-2 and QNLI) are also included as target tasks to provide comprehensive comparisons with prior parameter-efficient tuning methods, whose evaluations often focus on GLUE~\cite{lester-etal-2021-power,zaken2021bitfit}}.

\paragraph{Question answering.}
We use the MRQA 2019 shared task~\cite{fisch-etal-2019-mrqa} data to test on four large-scale QA datasets: Natural Questions~(NQ; \citealt{kwiatkowski-etal-2019-natural}), HotpotQA~(HQ;~\citealt{yang-etal-2018-hotpotqa}), NewsQA~(News;~\citealt{trischler-etal-2017-newsqa}) and SearchQA~(SQA; \citealt{dunn2017searchqa}).

\paragraph{Others.}
We experiments on four different datasets, whose tasks are related to the source tasks but domains differ. { SciTail}~\cite{Khot_Sabharwal_Clark_2018} is a scientific entailment dataset.
{Yelp-2}~\cite{zhang2015character} is a sentiment analysis dataset on Yelp reviews. 
{WinoGrande}~\cite{sakaguchi2020winogrande} is commonsense reasoning task in multiple choice format. {PAWS-Wiki}~\cite{zhang-etal-2019-paws} is a Wikipedia-based paraphrase detection dataset.

\begin{table*}[t!]
\centering
\footnotesize
\addtolength{\tabcolsep}{-4.25pt}  
\begin{tabular}{@{}l|a|ccccccccb|cccccb@{}}\toprule
\multicolumn{2}{c|}{} &\multicolumn{9}{c}{\textbf{GLUE}} & \multicolumn{6}{|c}{\textbf{Super GLUE}} \\ \midrule
\begin{tabular}[c]{@{}c@{}} data \\ (\# of train)\end{tabular}& \begin{tabular}[c]{@{}c@{}}param \\ / task\end{tabular} & \begin{tabular}[c]{@{}c@{}}MNLI\\ (393k)\end{tabular} & \begin{tabular}[c]{@{}c@{}}QQP\\ (364k)\end{tabular} & \begin{tabular}[c]{@{}c@{}}QNLI\\ (105k)\end{tabular} & \begin{tabular}[c]{@{}c@{}}SST-2\\ (67k)\end{tabular} & \begin{tabular}[c]{@{}c@{}}STS-B\\ (7k)\end{tabular} & \begin{tabular}[c]{@{}c@{}}MRPC\\ (3.7k)\end{tabular} & \begin{tabular}[c]{@{}c@{}}RTE\\ (2.5k)\end{tabular}  & \begin{tabular}[c]{@{}c@{}}CoLA\\ (8.5k) \end{tabular} & avg. & \begin{tabular}[c]{@{}c@{}}Multi\\ (5.1k) \end{tabular} & \begin{tabular}[c]{@{}c@{}}Bool\\ (9.4k) \end{tabular} & \begin{tabular}[c]{@{}c@{}}WiC\\ (6k) \end{tabular} & \begin{tabular}[c]{@{}c@{}}WSC\\ (554) \end{tabular} & \begin{tabular}[c]{@{}c@{}}CB\\ (250) \end{tabular} & avg.\\ \midrule
Fine-tuning      & 220M & {\bf 86.8} & {\bf 91.6} & 93.0&  {\bf 94.6}    & 89.7 & {\bf 90.2}   & 71.9 &  61.8 & 84.9 & 72.8 & 81.1    & {\bf 70.2} & 59.6 & { 85.7} & 73.9 \\
Adapter & 1.9M & 86.5& 90.2& {\bf 93.2} & 93.8    & { 90.7} & 85.3  & 71.9 & 64.0 & 84.5 & {\bf 75.9} & {\bf 82.5}   & 67.1 & { 67.3} & { 85.7} & {\bf 75.7} \\
{AdapterDrop} & 1.1M & 86.3 & 90.2 & 93.2 & 93.6& {\bf 91.4} & 86.3 & 71.2 & 62.7 & 84.4 &  72.9  & 82.3 & 68.3 & 67.3 &  85.7 & 75.3 \\
BitFit  & 280k &  85.3  &  90.1 & 93.0 & 94.2    & 90.9 & 86.8  & 67.6 & 58.2 & 83.3 & 74.5 & 79.6  & 70.0 & 59.6  & 78.6 & 72.5  \\
PT & 77k  & 81.3 & 89.7 &  92.8 &  90.9    & 89.5 & 68.1 & 54.7 & 10.6& 72.2 &  58.7 & 61.7   & 48.9  & 51.9  & 67.9 & 57.8  \\
SPoT  & 77k  & 85.4 & 90.1 & 93.0 & 93.4  & 90.0 & 79.7  & 69.8  &57.1  & 82.3  & 74.0 & 77.2   & 67.0  & 50.0   & 46.4 & 62.9 \\\midrule
Fine-tuning-m$^\dagger$ & 28M & 85.7 & 91.1 & 92.0 & 92.5 & 88.8 & {\bf 90.2} & 75.4 & 54.9 & 83.8 & -- & -- & -- & -- & -- & -- \\
Adapter-m$^\dagger$& {1.8M} & 86.3 & 90.5 & 93.2 & 93.0 & 89.9 & {\bf 90.2} & 70.3 & 61.5 & 84.4 & -- & -- & -- & -- & -- & --  \\
HyperFormer$^\dagger$ & 638k & 85.7 & 90.0 & 93.0 & 94.0 & 89.7 & 87.2 & 75.4 & 63.7 & 84.8 & -- & -- & -- & -- & -- & -- \\
HyperDecoder$^\ddagger$ & 1.8M &  86.0 & 90.5 & 93.4 & 94.0 & 90.5 & 87.7 & 71.7 & 55.9 & 83.7 & -- & -- & -- & -- & -- & -- \\
{AdapterFusion}$^{*}$ & -- &  84.2 & 90.7 & -- & 92.2 & -- & 90.3 & 76.8 & -- &  -- &  -- & 76.3 & --  &  -- & {\bf 92.1} & --    \\
\midrule
{\ours}    & 232k & 84.3 & 90.3 & 93.0 & 93.2 & 89.7 & 85.7  & { 73.4} & 57.4 & 83.4 &  74.4 & 78.8  & 66.8 & 53.8  & 78.6 & 70.5 \\ 
{\ours} -m  & 96k & 83.7 & 90.1 & {\bf 93.2} &  {94.3}  & {  90.8} & 87.3  & {\bf 82.7} & {\bf 64.3} & {\bf 85.8} & 74.4 & 78.5   & 66.5 & {\bf 69.2}  &  82.1 & 74.1 \\ 
\bottomrule
\end{tabular}
    \caption{Results on GLUE. {All of the results are based on T5-base models.} {For GLUE experiments, we exclude SQuAD and ReCoRD from source prompts inventories for comparison with prior work. We use Pearson Correlation for STS-B, F1 for MultiRC (Multi), and accuracy for other tasks as metrics. ``param/task'' denotes the number of the parameters trained for each task in GLUE. {$^\dagger$ from \citet{karimi-mahabadi-etal-2021-parameter_custom}; $^\ddagger$ from \citet{ivison2022hyperdecoders}; $^{*}$ from \citet{pfeiffer-etal-2021-adapterfusion} and their base LM is RoBERTa-base~\cite{liu2019roberta}. }
    }
    }
    \label{tab:main_results_glue}
\end{table*}
\begin{table*}[t]
    \centering
    \footnotesize
\addtolength{\tabcolsep}{-4pt}  
\begin{tabular}{@{}l|a|ccccb|ccccb@{}}
\toprule
\begin{tabular}[c]{@{}c@{}} data \\ (\# of train)\end{tabular} & \begin{tabular}[c]{@{}c@{}}params /\\ task\end{tabular} & \begin{tabular}[c]{@{}c@{}}NQ\\ (100k)\end{tabular} & \begin{tabular}[c]{@{}c@{}}HP\\ (72k)\end{tabular} & \begin{tabular}[c]{@{}c@{}}SQA\\ (117k)\end{tabular} & \begin{tabular}[c]{@{}c@{}}News\\ (74k)\end{tabular} & Avg.  & \begin{tabular}[c]{@{}c@{}}WG\\ (40k)\end{tabular} & \begin{tabular}[c]{@{}c@{}}Yelp\\ (100k)\end{tabular} & \begin{tabular}[c]{@{}c@{}}SciTail\\ (27k)\end{tabular} & \begin{tabular}[c]{@{}c@{}}PAWS\\ (49k)\end{tabular} & Avg.  \\ \midrule
Fine-tuning         & 220M & {\bf 75.1}  & 77.5 & 81.1 & 65.2 & {\bf 74.7} & {\bf 61.9} & 96.7  & {\bf 95.8}    & 94.1 & {\bf 87.1} \\
Adapter    & 1.9M & 74.2  & {\bf 77.6} & {\bf 81.4} & {\bf 65.6} & {\bf 74.7} & 59.2 & {\bf 96.9}  & 94.5    & {\bf 94.3} & 86.2 \\
BitFit     & 280k & 70.7  & 75.5 & 77.7 & 64.1 & 72.0 & 57.2 & 94.7  & 94.7    & 92.0 & 84.7 \\
Prompt tuning & 77k  & 67.9  & 72.9 & 75.7 & 61.1 & 69.4 & 49.6 & 95.1  & 87.9    & 55.8 & 72.1 \\
SPoT-t     & 77k  & 68.2  & 74.8 & 75.3 & 58.2 & 69.1 & 50.4 & 95.4 & 91.2    & 91.1 & 82.0 \\ 
\midrule
{\ours} & 232k & 70.4 & 75.2 & 77.3 & 62.8 & 71.4 & 57.6 & 96.7  & 93.1    & 92.1 & 84.9 \\ 
{\ours}-m & 134k & 72.5  & 76.7 & 78.0 & 63.9  & 72.8 & 58.6 & 96.2  & 94.6    & 92.8& 85.6 \\ 
\bottomrule
\end{tabular}
    \caption{Results on MRQA 2019 QA datasets, WinoGrande (WG), Yelp, Scitail and PAWS. 
    We use F1 for MRQA and accuracy for others. ``param/task'' denotes parameter trained per task in MRQA and others.} 
     \label{tab:main_mrqa}  

\end{table*}

\subsection{Baselines and Implementation Details}

\paragraph{Baselines.}
We compare \ours~
with: {\bf fine-tuning (FT)}; {\bf prompt tuning} ({\bf PT}; ~\citealt{lester-etal-2021-power}), where target prompt embeddings are initialized by randomly sampled top vocabularies; {\bf SPoT}~\cite{vu2021spot}, where target prompts are initialized by source prompt embeddings trained on other tasks ({details are in Appendix}); {\bf Adapter}~\cite{pmlr-v97-houlsby19a}, {\bf AdapterDrop}~\cite{ruckle-etal-2021-adapterdrop} and {\bf BitFit}~\cite{zaken2021bitfit}.  
{On GLUE, we also compare {\ours} with several state-of-the-art multi-task methods, which train a single model on different tasks: }
{\bf FT-multi-task (FT-m)}, {\bf Adapter-m}, {\bf HyperFormer}~\cite{karimi-mahabadi-etal-2021-parameter_custom}, {\bf HyperDecoder}~\cite{ivison2022hyperdecoders}, and {\bf AdapterFusion}~\cite{pfeiffer-etal-2021-adapterfusion}. 

\paragraph{Implementation details.} 
{
Although our methods, {\ours}~and {\ours}-m use the same six source task prompts, 
{\ours}-m trains a shared attention layer across multiple target tasks by conducting multi-task training, while \ours~trains a task-specific attention layer separately. }
Unless specified, we use T5-base as our base LMs for \ours~and all of the baselines.\footnote{Although the original prompt tuning paper uses T5 v1.1 LM-adapt as the backbone LMs, despite our extensive hyperparameter searches across five different learning rates and five different batch sizes, we could not reproduce the original results. We found that T5-LM adapt v1.1 was especially sensitive and hard to tune when we use it as a backbone LM for parameter-efficient approaches. Therefore, in this work we used T5 as backbone models. Prior work in this line also uses T5 as backbone models~\cite{mahabadi2021compacter}.
} 
If a dataset does not have public test split with annotations, we use a development set as our test set or split the development set into our development and test sets, following \citet{mahabadi2021compacter}. 
We train for 20 epochs on small datasets with less than 10k examples, 10 epochs on medium-size data with more than 10k examples, and 5 epochs on MRQA datasets and limit the maximum training data number of Yelp-2 to be 100k samples.
To make $\mathcal{G}$ learn a good prompt composition for efficient knowledge transfer, we introduce different learning rates for $\mathcal{G}$~\cite{ponti2022combining} and also pre-train and transfer the weights of $\mathcal{G}$ from the source tasks. 
More experimental details are in Appendix.

\paragraph{Prompt initialization.}
{
Each source prompt is initialized by randomly sampling tokens from the top vocabularies~ as in \citet{lester-etal-2021-power}. 
For target task prompt initialization, we use the MNLI source prompt for non-QA tasks and the SQuAD source prompt for QA, instead of initializing it with randomly sampled vocabularies for training stability.  }

\section{Results}
We present main results in \Sref{sec:main} and few-shot domain transfer experiments on sampled tasks in \Sref{sec:few_results}, demonstrating the effectiveness of ATTEMPT especially when the data is scarce. 
\Sref{sec:increase_base_size} further provides set of analyses. 

\subsection{Main Results}
\label{sec:main}
Tables \ref{tab:main_results_glue} and \ref{tab:main_mrqa} present the per-task performance of the GLUE and SuperGLUE datasets, and the other datasets, respectively.

\begin{figure}[t!]
\centering
\begin{subfigure}[t]{.38\linewidth}
  \centering
  \includegraphics[width=0.99\textwidth,keepaspectratio]{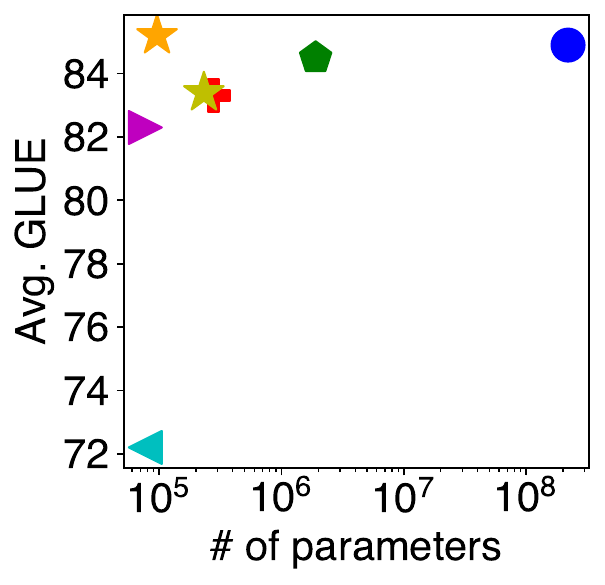}
   \captionsetup{width=0.99\textwidth}
  \caption{GLUE}
  \label{fig:glue_avg}
\end{subfigure}%
\begin{subfigure}[t]{.61\linewidth}
  \centering
  \includegraphics[width=\textwidth,keepaspectratio]{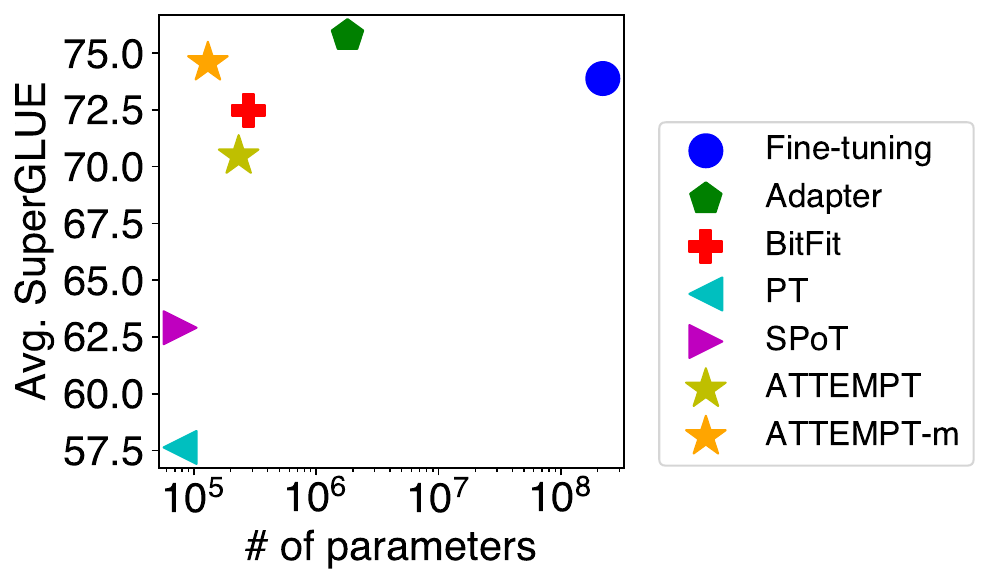} 
  \captionsetup{width=0.99\textwidth}
  \caption{SuperGLUE}
  \label{fig:superglue_avg}
\end{subfigure}%
\caption{Parameter-efficiency and average scores. We use T5-base for all of the models. }
\end{figure}
\paragraph{Performance vs. efficiency.} 
Figures \ref{fig:glue_avg} and \ref{fig:superglue_avg} compare the performance of different models versus their number of updated parameters on GLUE and SuperGLUE.  
\ours-m significantly outperforms PT, SPoT and BitFit by a large margin, and matches Adapter or Fine-tuning despite updating much fewer parameter per each task and keeping the LM completely frozen. \Tref{tab:main_results_glue} shows \ours~outperforms all of the multi-task baselines including recent HyperFormer or HyperDecoder. 
In addition to competitive performance on GLUE/SuperGLUE, \Tref{tab:main_mrqa} shows that \ours-m achieves 72.8 average F1 on MRQA, outperforming BitFit using twice as many parameters. 
Moreover, \ours-m yields 85.6\% average accuracy on WinoGrande, Yelp, SciTail and PAWS, outperforming BitFiT (84.7\%) and matching Adapter (86.2\%) that updates ten times more parameters.

\paragraph{ATTEMPT largely improves  prompt tuning.}
As pointed out by prior work~\cite{mahabadi2021compacter,lester-etal-2021-power,sung2022lst}, prompt tuning is sensitive to hyperparameters or initialization, and it has significantly lower performance on several datasets such as CoLA (10.2\%), BoolQ (61.7\%) or WiC (48.9\%).
SPoT~\cite{vu2021spot} improves the target task prompt initialization with a prompt trained on other related tasks, but it still under-performs other approaches, and requires searching the source tasks beforehand.  
\ours~largely outperforms those approaches on smaller datasets (e.g., CB, RTE), as well as large-scale MRQA datasets as shown in \Tref{tab:main_mrqa}.

\subsection{Few-shot Domain Adaptations}
\label{sec:few_results}
As shown in \Tref{tab:main_mrqa} ATTEMPT is particularly competitive on smaller dataset (e.g., RTE, WSC). 
{
Following~\citet{karimi-mahabadi-etal-2021-parameter_custom}, we conduct few-shot experiments on BoolQ, CB and SciTail, to further verify the effectiveness of \ours~under resource-constrained setup. 
Here, all of the models (Fine-tuning, Adapter, HyperFormer, SPoT and ATTEMPT) are first trained on the GLUE tasks and then transferred to new tasks using only $k$  ($k=4,16,32$) randomly sampled training data. More details of few-shot domain adaptation experiments are available at Appendix. 

Table~\ref{tab:few-shot} shows that
{\ours} significantly outperforms other methods in most of the setting. 
This indicate the effectiveness of transferring knowledge from multiple source tasks in a non-destructive manner in few-shot domain adaptation. 
}

\begin{table}
\centering
\footnotesize
\begin{tabular}{ll| c c cc |c}
\toprule

\multicolumn{2}{c|}{$k$-shot}& FT & AD & SPoT & HF & ATP \\ \midrule
 &4 & 50.5 & 53.5 & 50.5 & 48.0 & {\bf 61.8} \\
BoolQ & 16 & 56.5 & 51.4 & 50.6 & 50.2 & {\bf 60.0}  \\
& 32 & 58.4 & 54.5  & 61.2 & 58.3 & {\bf 65.3} \\
\midrule
& 4 & 57.8 & 51.1 &71.4 & 51.1 &  {\bf 82.1 }\\
CB & 16 & 77.0 & 74.8 & 64.3  & 74.8 & {\bf 78.5} \\
& 32 & 81.8 & 85.1 &64.3 & 81.5 & {\bf 85.7}  \\ \hline
& 4 & 79.6 & 79.5 &  69.6 &  {\bf 82.0} &  {80.2} \\
SciTail & 16 & 80.0 & { 83.3} & 71.9 & {\bf 86.6 }& 79.5 \\
& 32 & 82.0 & {85.1} & 71.9 & {\bf 85.9} & 80.2 \\ 
 \bottomrule
 \end{tabular}
 \caption{ Few-shot results ($k=\{4, 16, 32\}$). {FT, AD, HF and ATP denote Fine-tuning, Adapter, HyperFormer~\cite{karimi-mahabadi-etal-2021-parameter_custom} and {\ours}.}}
     \label{tab:few-shot}
 \end{table}

\subsection{Analyses}
\begin{figure}[t!]
\centering
\begin{subfigure}[t]{.33\linewidth}
  \centering
  \includegraphics[width=0.98\textwidth,keepaspectratio]{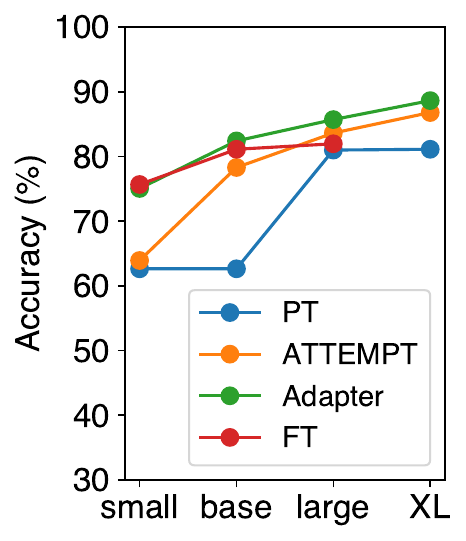}
   \captionsetup{width=0.95\textwidth}
  \caption{BoolQ}
  \label{fig:boolq}
\end{subfigure}%
\begin{subfigure}[t]{.33\linewidth}
  \centering
  \includegraphics[width=0.98\textwidth,keepaspectratio]{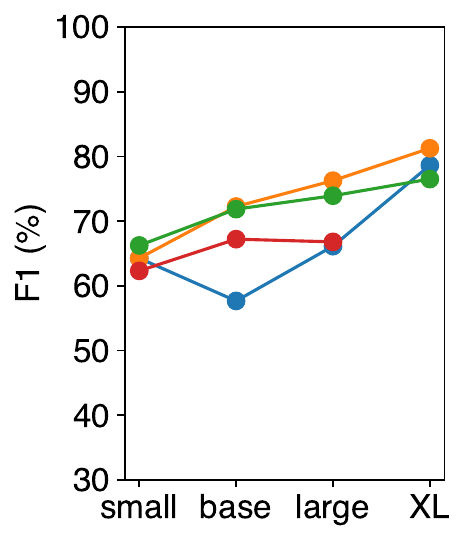} 
  \captionsetup{width=0.95\textwidth}
  \caption{MultiRC}
  \label{fig:multirc}
\end{subfigure}%
\begin{subfigure}[t]{.33\linewidth}
  \centering
  \includegraphics[width=0.98\textwidth,keepaspectratio]{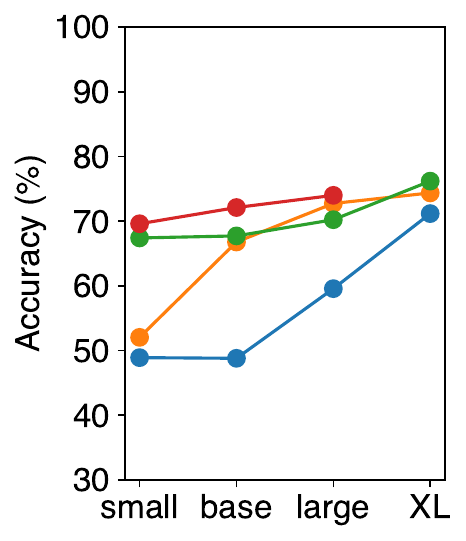}
   \captionsetup{width=0.95\textwidth}
  \caption{WiC}
  \label{fig:wic}
\end{subfigure}%
\caption{Performance with different backbone LMs. }
\label{fig:powerofscale}
\end{figure}

\paragraph{Power of scale.} 
\label{sec:increase_base_size}
We empirically analyze how increasing the backbone LM size affects \ours~performance.  
Figure~\ref{fig:powerofscale} summarizes the performance of  Adapter, \ours, prompt tuning (PT), and fully fine-tuned (FT) models vs. LM sizes on three SuperGLUE datasets.\footnote{We could not finetune T5-XL under the same computational constraints as the other baselines, so we do not report the fine-tuning with T5-XL performance. } 
\ours~largely benefits from backbone LM size increase. This is aligned with the finding of \citet{lester-etal-2021-power} that show  prompt tuning is particularly effective when the backbone LM is larger. Moreover, \ours~matches fully fine-tuned models even with T5-base or T5-large. This is in contrast to prompt tuning methods that suffers when the backbone LM is smaller.
Furthermore, \ours~performs on par with or outperforming Adapter with T5-3B, while updating 37 times less parameters.

\paragraph{Ablation studies.}
\label{sec:ablation}
We compare different variants of \ours~to see the effect of each of the design choices.\footnote{The ablations are mainly conducted over BoolQ, NewsQA and WinoGrande.}
We ablate \ours~with (a) {\it no target}, which neither initializes nor adds target task prompts in \Eref{eq:final_soft}, to assess the feasibility of adapting to a new task by only interpolating pre-trained source prompts; (b) {\it no attention}, which gives constant score $a_j = 1 / t$ to all source prompts in \Eref{eq:soft_max}, discarding attentions; (c) {\it single prompt}, which uses only a single source prompt to assess the effect of transferring knowledge from multiple tasks.
{Single prompt ablation is similar to SPoT except that instead of using source prompts for initialization and updating the source prompt parameters during training, we keep the source prompt intact while updating the target task prompt and interpolate the source and target prompts. }
\begin{table}[t!]
\small
    \centering
    \begin{tabular}{l| ccc}
\toprule
&  BoolQ & NewsQA & WG \\\midrule
{\ours}-m &   78.29 & 61.58 &58.57 \\
{\ours} & 77.06  & 61.84 & 57.61  \\\hline
no target & 50.89  & 55.26 & 47.89   \\
no attention &  73.57  & 52.55 & 56.03  \\
single prompt & 76.25   & 60.92 & 55.56  \\
 \bottomrule
 \end{tabular}
    \caption{Results of ablation studies. ``WG'' denotes WinoGrande. For NewsQA ablation, we used randomly sampled 10k data for training for quick ablation. 
    }
    \label{tab:main_ablation}
\end{table}

\Tref{tab:main_ablation} indicates that all components contribute to performance improvements. 
Adding a trainable target-task-specific prompt (no target) is crucial to achieve good performance on all of the datasets, especially on BoolQ and WinoGrande. 
Constant attention causes large performance drop,  especially on BoolQ and NewsQA, indicating that it is important to have learned attentions rather than simply averaging the multiple source prompts. 
{
Although the single prompt ablation baseline  outperforms SPoT, possibly due to the non-destructive soft prompt transfer of \ours,
there is notable performance decline relative to \ours. 
This demonstrates the effectiveness of leveraging multiple soft prompts to transfer knowledge from multiple diverse tasks. }

\begin{figure}[t!]
\centering
\begin{subfigure}[t]{.49\linewidth}
  \centering
  \includegraphics[width=0.9\textwidth,keepaspectratio]{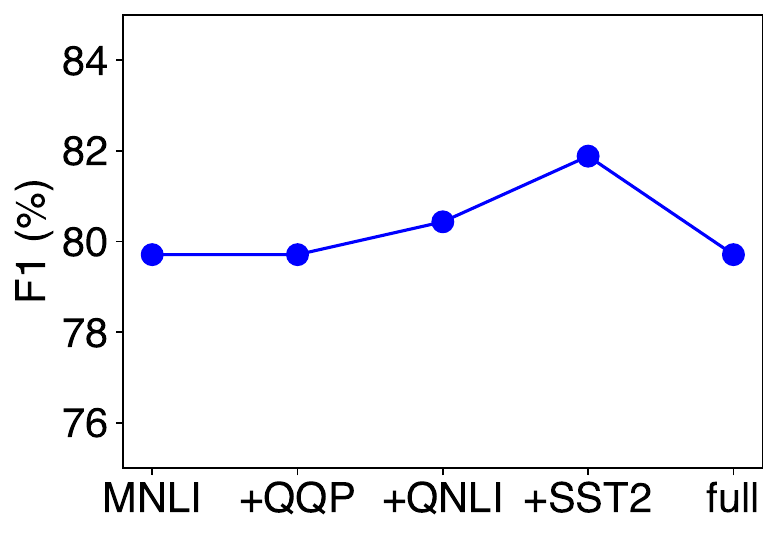}
  \captionsetup{width=0.95\textwidth}
  \caption{RTE}
  \label{fig:modul_rte}
\end{subfigure}%
\begin{subfigure}[t]{.49\linewidth}
  \centering
  \includegraphics[width=0.9\textwidth,keepaspectratio]{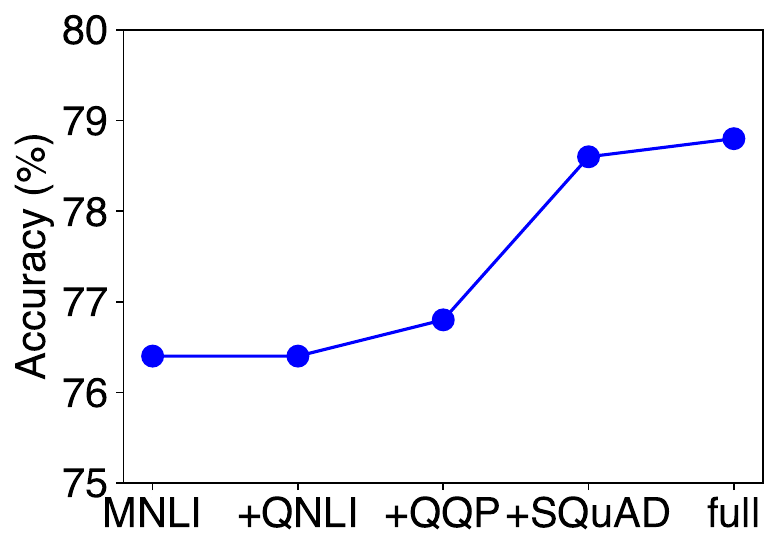}
  \captionsetup{width=0.95\textwidth}
  \caption{BoolQ}
  \label{fig:module_bool}
\end{subfigure}%
\caption{Performance on RTE and BoolQ dev sets when source prompts are added one by one, starting from the MNLI source prompt only.
}
\label{fig:modular}
\end{figure}
\begin{figure}[t!]
\includegraphics[width=7.5cm]{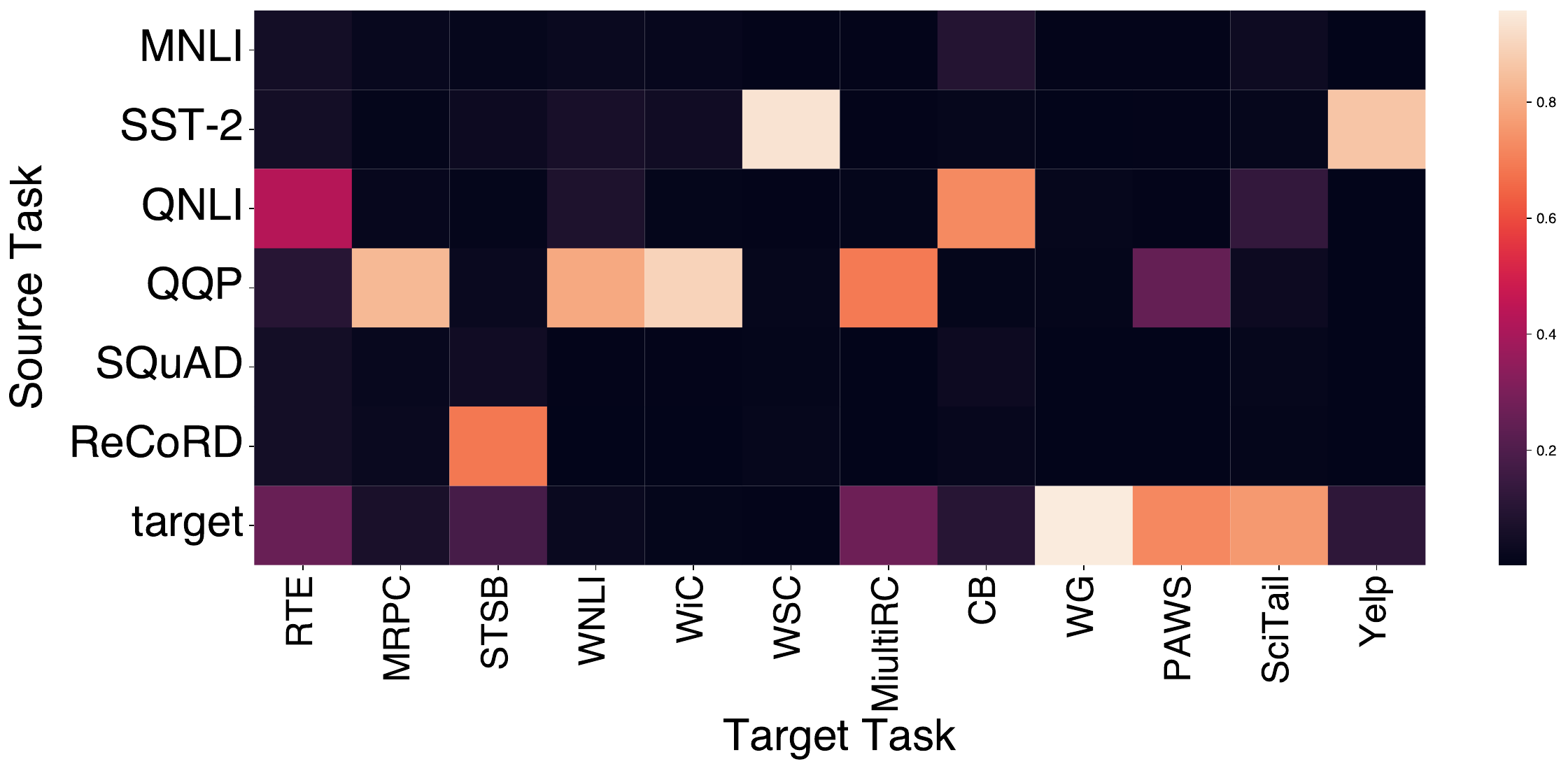}\caption{Attention visualizations of \ours. }\label{fig:attn_vis}
\end{figure}

\paragraph{Modularity: effects of variable source prompts.}
\label{sec:variable_source_prompt}
We study the modular nature of \ours~that enables flexibly adding or removing source tasks. 
Figure~\ref{fig:modular} shows how including source tasks affects  the final performance of \ours~on two benchmarks, BoolQ and RTE. On both of the datasets, adding more source task prompts gives performance improvements, with an exception of adding SQuAD and ReCoRD on RTE (``full'' in \Fref{fig:modul_rte}). 
This potentially happens because of the negative transfer due to the different natures of QA and RTE, while adding the two QA source prompts helps in BoolQ. 
This indicates the effectiveness of \ours's modularity, where one can flexibly remove or add new tasks to build a powerful multi-task model.

\paragraph{Interpretability: analysis on attentions. }
\Fref{fig:attn_vis} shows the attention weight matrix between source and target tasks by \ours. 
{Note that for the target task prompt, we present the $a_{t+1}$ weight before adding 1.}
Attention patterns differ for different tasks.
{Generally, $\mathcal{G}$ gives higher attentions to related source tasks: Yelp $\rightarrow$ SST-2, or PAWS-Wiki $\rightarrow$ QQP, which are the same tasks but are different in domains.}
QQP is often highly attended by some tasks that are seemingly different from paraphrasing (e.g., MultiRC, WNLI), which may indicate underlying task similarities between those tasks. 
{Unlike the underlying task similarities, MNLI is not highly attended by some highly-related target tasks such as RTE. We hypothesize that this is because the target task prompts for those tasks are initialized with the MNLI source prompt, and thus \ours~may try to attend to other tasks. }
On WinoGrande or SciTail, $\mathcal{G}$ gives large attentions to the target task embeddings (``target''); this may be because those two tasks have significantly different task format or input domain, and $\mathcal{G}$ ignores source prompts more.   

\section{ Related Work}
\paragraph{Parameter-efficient tuning.}
Here, we enlist additional parameter-efficient tuning methods that are close to our work.
AdapterFusion~\cite{pfeiffer-etal-2021-adapterfusion} compose multiple different adapters by learning task-specific compositions on each task, and \citet{friedman-etal-2021-single} take an average of multiple adapter layers after training adapters individually on different QA datasets. 
HyperFormer~\cite{karimi-mahabadi-etal-2021-parameter_custom} and HyperDecoder~\cite{ivison2022hyperdecoders} train a shared hyper network to generate parameters of adapter layers.
\citet{qin-eisner-2021-learning} introduce mixture of soft prompts, where predictions given different prompts are ensembled for the same knowledge base relationship types. 
IDPG~\cite{wu2022idpg} and Instance-Dependent Prompt Tuning~\cite{levine2022standing} learn to generate instance-wise prompts given the input encoded by LMs.  
Compared to the previous work, our main focus is transferring knowledge from multiple tasks to produce soft prompts rather than learning to generate them from scratch, and is much more efficient in terms of parameters and inference time. 

Concurrent to our work, \citet{liu2022few} introduce (IA)$^3$ that multiplies intermediate activation by learned vectors for few-shot learning. 
{\citet{wang2022learning} shows that combining a set of prompts retrieved from the prompt pool by a key-value mechanism yields competitive performance in computer vision continual learning. }
For generation tasks, \citet{li2022learning} transfer multiple source prompts  using multi-key memory network for prompt clustering and multi-head attention taking another LM output. 
{In contrast, we present an efficient multi-task tuning that is effective in diverse NLP tasks. More importantly, prior work often relies on priors such as pre-computed clusters or another LM's predictions of which tasks should be used as source tasks. 
\ours~removes the necessity of such priors by training an attention layer that learn to focus on relevant source tasks.}

{Several recent lines of research attempt to adapt a massive multi-task LM trained with instructions or demonstrations to a new task without any parameter updates~\cite{sanh2022multitask,min2021metaicl,wang2022benchmarking,wei2022finetuned}. 
The main focus of this paper is how to efficiently transfer rich multi-task knowledge from source tasks to target tasks with training data during target task training, while those work often emphasize on zero or few-shot transfer without any parameter updates.}

\paragraph{Modular multi-task training.}
There is a large literature on composing multiple separate networks to handle different sub-tasks~\cite{jacobs1991adaptive,jacobs1991task,andreas2016neural,mccann2018natural}. 
As the LM size expands, several recent work tries to sparsely activate or employ light-weight modules for efficient multi-task learning~\cite{gupta2022sparsely,ponti2022combining,fedus2021switch}. 
In particular, we share the same intuition as the 
concurrent work~\cite{ponti2022combining}, which combines several skills encapsulated in parameter-efficient modules; however, our main focus is on how to transfer and share knowledge from resource-rich tasks in a super parameter-efficient way, while they focus on improving few-shot generalization ability. Moreover, \ours~keeps LMs intact and updates fewer parameters.

\section{Conclusion}
We present a new parameter-effluent tuning method \ours, which learns to produce instance-wise prompts  by interpolating multiple reusable soft prompts trained on source tasks and a new task-specific prompt, while keeping the original LM frozen. Our large-scale experiments demonstrate that {\ours} achieves a great trade-off between task performance and efficiency, introducing an interpretable and modular task transfer. 

\section*{Limitations}
Despite its parameter-efficiency and strong empirical results, {\ours} has several limitations: 
{
First, as prompt tuning increases the input token length by $m$ prompt tokens, it increases the memory footprint and computational costs~\cite{mahabadi2021compacter}, although \citet{lester-etal-2021-power} found that prompt length can be shortened when larger LMs are used as backbone models. 
We investigate this issue in Appendix Section~\ref{sec:memory_footprint}.}
Secondly, as the first step toward multi-task knowledge transfer via soft prompts, our evaluation focuses on classification and QA tasks, and our target tasks do not include the tasks that require long sequence generations (e.g., summarization). 
Future work can explore applications of \ours~to more diverse sets of tasks. 
{
In addition, we use representative six NLP tasks as source tasks, but do not explore a large-scale experiments on many source task combinations. 
We will release pretrained source prompts and easily extendable code to facilitate future work on multi-task transfer via soft prompt transfer. 
}
Lastly, we do not test {\ours} on non-English tasks, and we will investigate the effectiveness of {\ours} in non-English languages or apply {\ours} for cross-lingual transfer. 

\section*{Ethics Statement}
{\ours} is trying to improve parameter-efficiency and transferability of models so that groups with limited computational resources can still get benefit from state-of-the-art large-scale models.  
All of the experiments are based on widely-used general purpose datasets, which are unlikely to include harmful content. 
However, several datasets such as Yelp Review are created from existing review sites, and may have more risks of privacy issues or harmful content than some other datasets based on news or encyclopedic websites.

\section*{Acknowledgement}
This research was supported by NSF IIS-2044660, ONR N00014-18-1-2826, a Sloan fellowship and gifts from AI2, and the Nakajima Foundation Fellowship. 
We thank UW NLP and Allen NLP group members for their insightful discussion and Hamish Ivison, Sandy Kaplan, Sewon Min, Ofir Press, Yizhong Wang, and the EMNLP 2022 anonymous reviewers for their helpful
feedback on this paper. 

\bibliography{custom}

\begin{thebibliography}{79}
\expandafter\ifx\csname natexlab\endcsname\relax\def\natexlab#1{#1}\fi

\bibitem[{Aghajanyan et~al.(2021)Aghajanyan, Gupta, Shrivastava, Chen,
  Zettlemoyer, and Gupta}]{aghajanyan-etal-2021-muppet}
Armen Aghajanyan, Anchit Gupta, Akshat Shrivastava, Xilun Chen, Luke
  Zettlemoyer, and Sonal Gupta. 2021.
\newblock \href {https://aclanthology.org/2021.emnlp-main.468} {Muppet: Massive
  multi-task representations with pre-finetuning}.
\newblock In \emph{EMNLP}.

\bibitem[{Andreas et~al.(2016)Andreas, Rohrbach, Darrell, and
  Klein}]{andreas2016neural}
Jacob Andreas, Marcus Rohrbach, Trevor Darrell, and Dan Klein. 2016.
\newblock \href {https://arxiv.org/abs/1511.02799} {Neural module networks}.
\newblock In \emph{CVPR}.

\bibitem[{Aribandi et~al.(2022)Aribandi, Tay, Schuster, Rao, Zheng, Mehta,
  Zhuang, Tran, Bahri, Ni, Gupta, Hui, Ruder, and Metzler}]{aribandi2021ext5}
Vamsi Aribandi, Yi~Tay, Tal Schuster, Jinfeng Rao, Huaixiu~Steven Zheng,
  Sanket~Vaibhav Mehta, Honglei Zhuang, Vinh~Q. Tran, Dara Bahri, Jianmo Ni,
  Jai Gupta, Kai Hui, Sebastian Ruder, and Donald Metzler. 2022.
\newblock \href {https://openreview.net/forum?id=Vzh1BFUCiIX} {Ext5: Towards
  extreme multi-task scaling for transfer learning}.
\newblock In \emph{ICLR}.

\bibitem[{Ba et~al.(2016)Ba, Kiros, and Hinton}]{ba2016layer}
Jimmy~Lei Ba, Jamie~Ryan Kiros, and Geoffrey~E Hinton. 2016.
\newblock \href {https://arxiv.org/abs/1607.06450} {Layer normalization}.
\newblock \emph{arXiv preprint arXiv:1607.06450}.

\bibitem[{Ben~Zaken et~al.(2022)Ben~Zaken, Goldberg, and
  Ravfogel}]{zaken2021bitfit}
Elad Ben~Zaken, Yoav Goldberg, and Shauli Ravfogel. 2022.
\newblock \href {https://aclanthology.org/2022.acl-short.1} {{B}it{F}it: Simple
  parameter-efficient fine-tuning for transformer-based masked
  language-models}.
\newblock In \emph{ACL}.

\bibitem[{Brown et~al.(2020)Brown, Mann, Ryder, Subbiah, Kaplan, Dhariwal,
  Neelakantan, Shyam, Sastry, Askell et~al.}]{brown2020language}
Tom Brown, Benjamin Mann, Nick Ryder, Melanie Subbiah, Jared~D Kaplan, Prafulla
  Dhariwal, Arvind Neelakantan, Pranav Shyam, Girish Sastry, Amanda Askell,
  et~al. 2020.
\newblock \href
  {https://papers.nips.cc/paper/2020/hash/1457c0d6bfcb4967418bfb8ac142f64a-Abstract.html}
  {Language models are few-shot learners}.
\newblock In \emph{NeurIPS}.

\bibitem[{Cer et~al.(2017)Cer, Diab, Agirre, Lopez-Gazpio, and
  Specia}]{cer-etal-2017-semeval}
Daniel Cer, Mona Diab, Eneko Agirre, I{\~n}igo Lopez-Gazpio, and Lucia Specia.
  2017.
\newblock \href {https://aclanthology.org/S17-2001} {{S}em{E}val-2017 task 1:
  Semantic textual similarity multilingual and crosslingual focused
  evaluation}.
\newblock In \emph{Proceedings of the 11th International Workshop on Semantic
  Evaluation ({S}em{E}val-2017)}.

\bibitem[{Chowdhery et~al.(2022)Chowdhery, Narang, Devlin, Bosma, Mishra,
  Roberts, Barham, Chung, Sutton, Gehrmann et~al.}]{chowdhery2022palm}
Aakanksha Chowdhery, Sharan Narang, Jacob Devlin, Maarten Bosma, Gaurav Mishra,
  Adam Roberts, Paul Barham, Hyung~Won Chung, Charles Sutton, Sebastian
  Gehrmann, et~al. 2022.
\newblock \href {https://arxiv.org/abs/2204.02311} {Pa{LM}: Scaling language
  modeling with pathways}.
\newblock \emph{arXiv preprint arXiv:2204.02311}.

\bibitem[{Clark et~al.(2019)Clark, Lee, Chang, Kwiatkowski, Collins, and
  Toutanova}]{clark-etal-2019-boolq}
Christopher Clark, Kenton Lee, Ming-Wei Chang, Tom Kwiatkowski, Michael
  Collins, and Kristina Toutanova. 2019.
\newblock \href {https://aclanthology.org/N19-1300} {{B}ool{Q}: Exploring the
  surprising difficulty of natural yes/no questions}.
\newblock In \emph{NAACL}.

\bibitem[{De~Marneffe et~al.(2019)De~Marneffe, Simons, and
  Tonhauser}]{de2019commitmentbank}
Marie-Catherine De~Marneffe, Mandy Simons, and Judith Tonhauser. 2019.
\newblock \href
  {https://ojs.ub.uni-konstanz.de/sub/index.php/sub/article/view/601} {The
  commitmentbank: Investigating projection in naturally occurring discourse}.
\newblock In \emph{Sinn und Bedeutung 23}.

\bibitem[{Demszky et~al.(2018)Demszky, Guu, and
  Liang}]{demszky2018transforming}
Dorottya Demszky, Kelvin Guu, and Percy Liang. 2018.
\newblock \href {https://arxiv.org/abs/1809.02922} {Transforming question
  answering datasets into natural language inference datasets}.
\newblock \emph{arXiv preprint arXiv:1809.02922}.

\bibitem[{Devlin et~al.(2019)Devlin, Chang, Lee, and
  Toutanova}]{devlin-etal-2019-bert}
Jacob Devlin, Ming-Wei Chang, Kenton Lee, and Kristina Toutanova. 2019.
\newblock \href {https://aclanthology.org/N19-1423} {{BERT}: Pre-training of
  deep bidirectional transformers for language understanding}.
\newblock In \emph{NAACL}.

\bibitem[{Dodge et~al.(2020)Dodge, Ilharco, Schwartz, Farhadi, Hajishirzi, and
  Smith}]{dodge2020fine}
Jesse Dodge, Gabriel Ilharco, Roy Schwartz, Ali Farhadi, Hannaneh Hajishirzi,
  and Noah Smith. 2020.
\newblock \href {https://arxiv.org/abs/2002.06305} {Fine-tuning pretrained
  language models: Weight initializations, data orders, and early stopping}.
\newblock \emph{arXiv preprint arXiv:2002.06305}.

\bibitem[{Dolan and Brockett(2005)}]{dolan-brockett-2005-automatically}
William~B. Dolan and Chris Brockett. 2005.
\newblock \href {https://aclanthology.org/I05-5002} {Automatically constructing
  a corpus of sentential paraphrases}.
\newblock In \emph{Proceedings of the Third International Workshop on
  Paraphrasing ({IWP})}.

\bibitem[{Dunn et~al.(2017)Dunn, Sagun, Higgins, Guney, Cirik, and
  Cho}]{dunn2017searchqa}
Matthew Dunn, Levent Sagun, Mike Higgins, V~Ugur Guney, Volkan Cirik, and
  Kyunghyun Cho. 2017.
\newblock \href {https://arxiv.org/abs/1704.05179} {Search{QA}: A new q\&a
  dataset augmented with context from a search engine}.
\newblock \emph{arXiv preprint arXiv:1704.05179}.

\bibitem[{Elfwing et~al.(2017)Elfwing, Uchibe, and Doya}]{elfwing2017sigmoid}
Stefan Elfwing, Eiji Uchibe, and Kenji Doya. 2017.
\newblock \href {https://arxiv.org/abs/1702.03118} {Sigmoid-weighted linear
  units for neural network function approximation in reinforcement learning}.
\newblock \emph{arXiv preprint arXiv:1702.03118}.

\bibitem[{Fedus et~al.(2022)Fedus, Zoph, and Shazeer}]{fedus2021switch}
William Fedus, Barret Zoph, and Noam Shazeer. 2022.
\newblock \href {http://jmlr.org/papers/v23/21-0998.html} {Switch transformers:
  Scaling to trillion parameter models with simple and efficient sparsity}.
\newblock \emph{JMLR}.

\bibitem[{Fisch et~al.(2019)Fisch, Talmor, Jia, Seo, Choi, and
  Chen}]{fisch-etal-2019-mrqa}
Adam Fisch, Alon Talmor, Robin Jia, Minjoon Seo, Eunsol Choi, and Danqi Chen.
  2019.
\newblock \href {https://aclanthology.org/D19-5801} {{MRQA} 2019 shared task:
  Evaluating generalization in reading comprehension}.
\newblock In \emph{Proceedings of the 2nd Workshop on Machine Reading for
  Question Answering}.

\bibitem[{Friedman et~al.(2021)Friedman, Dodge, and
  Chen}]{friedman-etal-2021-single}
Dan Friedman, Ben Dodge, and Danqi Chen. 2021.
\newblock \href {https://aclanthology.org/2021.emnlp-main.495} {Single-dataset
  experts for multi-dataset question answering}.
\newblock In \emph{EMNLP}.

\bibitem[{Giampiccolo et~al.(2007)Giampiccolo, Magnini, Dagan, and
  Dolan}]{giampiccolo-etal-2007-third}
Danilo Giampiccolo, Bernardo Magnini, Ido Dagan, and Bill Dolan. 2007.
\newblock \href {https://aclanthology.org/W07-1401} {The third {PASCAL}
  recognizing textual entailment challenge}.
\newblock In \emph{Proceedings of the {ACL}-{PASCAL} Workshop on Textual
  Entailment and Paraphrasing}.

\bibitem[{Gupta et~al.(2022)Gupta, Mukherjee, Subudhi, Gonzalez, Jose,
  Awadallah, and Gao}]{gupta2022sparsely}
Shashank Gupta, Subhabrata Mukherjee, Krishan Subudhi, Eduardo Gonzalez, Damien
  Jose, Ahmed~H Awadallah, and Jianfeng Gao. 2022.
\newblock \href {https://arxiv.org/abs/2204.07689} {Sparsely activated
  mixture-of-experts are robust multi-task learners}.
\newblock \emph{arXiv preprint arXiv:2204.07689}.

\bibitem[{He et~al.(2022)He, Zhou, Ma, Berg-Kirkpatrick, and
  Neubig}]{he2021towards}
Junxian He, Chunting Zhou, Xuezhe Ma, Taylor Berg-Kirkpatrick, and Graham
  Neubig. 2022.
\newblock \href {https://openreview.net/forum?id=0RDcd5Axok} {Towards a unified
  view of parameter-efficient transfer learning}.
\newblock In \emph{ICLR}.

\bibitem[{Hendrycks and Gimpel(2016)}]{hendrycks2016gaussian}
Dan Hendrycks and Kevin Gimpel. 2016.
\newblock \href {https://arxiv.org/abs/1606.08415} {Gaussian error linear units
  ({GELUs})}.
\newblock \emph{arXiv preprint arXiv:1606.08415}.

\bibitem[{Houlsby et~al.(2019)Houlsby, Giurgiu, Jastrzebski, Morrone,
  De~Laroussilhe, Gesmundo, Attariyan, and Gelly}]{pmlr-v97-houlsby19a}
Neil Houlsby, Andrei Giurgiu, Stanislaw Jastrzebski, Bruna Morrone, Quentin
  De~Laroussilhe, Andrea Gesmundo, Mona Attariyan, and Sylvain Gelly. 2019.
\newblock \href {https://proceedings.mlr.press/v97/houlsby19a.html}
  {Parameter-efficient transfer learning for {NLP}}.
\newblock In \emph{ICML}.

\bibitem[{Ivison and Peters(2022)}]{ivison2022hyperdecoders}
Hamish Ivison and Matthew~E Peters. 2022.
\newblock \href {https://arxiv.org/abs/2203.08304} {Hyperdecoders:
  Instance-specific decoders for multi-task {NLP}}.
\newblock \emph{arXiv preprint arXiv:2203.08304}.

\bibitem[{Jacobs et~al.(1991{\natexlab{a}})Jacobs, Jordan, and
  Barto}]{jacobs1991task}
Robert~A Jacobs, Michael~I Jordan, and Andrew~G Barto. 1991{\natexlab{a}}.
\newblock \href
  {https://onlinelibrary.wiley.com/doi/abs/10.1207/s15516709cog1502_2} {Task
  decomposition through competition in a modular connectionist architecture:
  The what and where vision tasks}.
\newblock \emph{Cognitive science}.

\bibitem[{Jacobs et~al.(1991{\natexlab{b}})Jacobs, Jordan, Nowlan, and
  Hinton}]{jacobs1991adaptive}
Robert~A Jacobs, Michael~I Jordan, Steven~J Nowlan, and Geoffrey~E Hinton.
  1991{\natexlab{b}}.
\newblock \href {https://ieeexplore.ieee.org/document/6797059} {Adaptive
  mixtures of local experts}.
\newblock \emph{Neural computation}.

\bibitem[{Khashabi et~al.(2018)Khashabi, Chaturvedi, Roth, Upadhyay, and
  Roth}]{khashabi-etal-2018-looking}
Daniel Khashabi, Snigdha Chaturvedi, Michael Roth, Shyam Upadhyay, and Dan
  Roth. 2018.
\newblock \href {https://aclanthology.org/N18-1023} {Looking beyond the
  surface: A challenge set for reading comprehension over multiple sentences}.
\newblock In \emph{NAACL}.

\bibitem[{Khashabi et~al.(2022)Khashabi, Lyu, Min, Qin, Richardson, Welleck,
  Hajishirzi, Khot, Sabharwal, Singh, and Choi}]{khashabi2021prompt}
Daniel Khashabi, Xinxi Lyu, Sewon Min, Lianhui Qin, Kyle Richardson, Sean
  Welleck, Hannaneh Hajishirzi, Tushar Khot, Ashish Sabharwal, Sameer Singh,
  and Yejin Choi. 2022.
\newblock \href {https://aclanthology.org/2022.naacl-main.266} {Prompt
  waywardness: The curious case of discretized interpretation of continuous
  prompts}.
\newblock In \emph{NAACL}.

\bibitem[{Khashabi et~al.(2020)Khashabi, Min, Khot, Sabharwal, Tafjord, Clark,
  and Hajishirzi}]{khashabi-etal-2020-unifiedqa}
Daniel Khashabi, Sewon Min, Tushar Khot, Ashish Sabharwal, Oyvind Tafjord,
  Peter Clark, and Hannaneh Hajishirzi. 2020.
\newblock \href {https://aclanthology.org/2020.findings-emnlp.171}
  {{UNIFIEDQA}: Crossing format boundaries with a single {QA} system}.
\newblock In \emph{EMNLP}.

\bibitem[{Khot et~al.(2018)Khot, Sabharwal, and
  Clark}]{Khot_Sabharwal_Clark_2018}
Tushar Khot, Ashish Sabharwal, and Peter Clark. 2018.
\newblock \href {https://ojs.aaai.org/index.php/AAAI/article/view/12022}
  {Scitail: A textual entailment dataset from science question answering}.
\newblock In \emph{AAAI}.

\bibitem[{Kingma and Ba(2015)}]{kingma2014adam}
Diederik~P Kingma and Jimmy Ba. 2015.
\newblock \href {https://arxiv.org/abs/1412.6980} {Adam: A method for
  stochastic optimization}.
\newblock In \emph{ICLR}.

\bibitem[{Kwiatkowski et~al.(2019)Kwiatkowski, Palomaki, Redfield, Collins,
  Parikh, Alberti, Epstein, Polosukhin, Devlin, Lee, Toutanova, Jones, Kelcey,
  Chang, Dai, Uszkoreit, Le, and Petrov}]{kwiatkowski-etal-2019-natural}
Tom Kwiatkowski, Jennimaria Palomaki, Olivia Redfield, Michael Collins, Ankur
  Parikh, Chris Alberti, Danielle Epstein, Illia Polosukhin, Jacob Devlin,
  Kenton Lee, Kristina Toutanova, Llion Jones, Matthew Kelcey, Ming-Wei Chang,
  Andrew~M. Dai, Jakob Uszkoreit, Quoc Le, and Slav Petrov. 2019.
\newblock \href {https://aclanthology.org/Q19-1026} {Natural questions: A
  benchmark for question answering research}.
\newblock \emph{TACL}.

\bibitem[{Lester et~al.(2021)Lester, Al-Rfou, and
  Constant}]{lester-etal-2021-power}
Brian Lester, Rami Al-Rfou, and Noah Constant. 2021.
\newblock \href {https://aclanthology.org/2021.emnlp-main.243} {The power of
  scale for parameter-efficient prompt tuning}.
\newblock In \emph{EMNLP}.

\bibitem[{Levesque et~al.(2012)Levesque, Davis, and
  Morgenstern}]{10.5555/3031843.3031909}
Hector~J. Levesque, Ernest Davis, and Leora Morgenstern. 2012.
\newblock \href {http://www.cs.nyu.edu/faculty/davise/papers/WSKR2012.pdf} {The
  winograd schema challenge}.
\newblock In \emph{Proceedings of the Thirteenth International Conference on
  Principles of Knowledge Representation and Reasoning}.

\bibitem[{Levine et~al.(2022)Levine, Dalmedigos, Ram, Zeldes, Jannai, Muhlgay,
  Osin, Lieber, Lenz, Shalev-Shwartz et~al.}]{levine2022standing}
Yoav Levine, Itay Dalmedigos, Ori Ram, Yoel Zeldes, Daniel Jannai, Dor Muhlgay,
  Yoni Osin, Opher Lieber, Barak Lenz, Shai Shalev-Shwartz, et~al. 2022.
\newblock \href {https://arxiv.org/abs/2204.10019} {Standing on the shoulders
  of giant frozen language models}.
\newblock \emph{arXiv preprint arXiv:2204.10019}.

\bibitem[{Li et~al.(2022)Li, Tang, Nie, Wen, and Zhao}]{li2022learning}
Junyi Li, Tianyi Tang, Jian-Yun Nie, Ji-Rong Wen, and Wayne~Xin Zhao. 2022.
\newblock \href {https://arxiv.org/abs/2205.01543} {Learning to transfer
  prompts for text generation}.
\newblock In \emph{NAACL}.

\bibitem[{Li and Liang(2021)}]{li-liang-2021-prefix}
Xiang~Lisa Li and Percy Liang. 2021.
\newblock \href {https://aclanthology.org/2021.acl-long.353} {Prefix-tuning:
  Optimizing continuous prompts for generation}.
\newblock In \emph{ACL}.

\bibitem[{Liu et~al.(2022)Liu, Tam, Muqeeth, Mohta, Huang, Bansal, and
  Raffel}]{liu2022few}
Haokun Liu, Derek Tam, Mohammed Muqeeth, Jay Mohta, Tenghao Huang, Mohit
  Bansal, and Colin Raffel. 2022.
\newblock \href {https://arxiv.org/abs/2205.05638} {Few-shot
  parameter-efficient fine-tuning is better and cheaper than in-context
  learning}.
\newblock \emph{arXiv preprint arXiv:2205.05638}.

\bibitem[{Liu et~al.(2019{\natexlab{a}})Liu, He, Chen, and
  Gao}]{liu-etal-2019-multi}
Xiaodong Liu, Pengcheng He, Weizhu Chen, and Jianfeng Gao. 2019{\natexlab{a}}.
\newblock \href {https://aclanthology.org/P19-1441} {Multi-task deep neural
  networks for natural language understanding}.
\newblock In \emph{ACL}.

\bibitem[{Liu et~al.(2019{\natexlab{b}})Liu, Ott, Goyal, Du, Joshi, Chen, Levy,
  Lewis, Zettlemoyer, and Stoyanov}]{liu2019roberta}
Yinhan Liu, Myle Ott, Naman Goyal, Jingfei Du, Mandar Joshi, Danqi Chen, Omer
  Levy, Mike Lewis, Luke Zettlemoyer, and Veselin Stoyanov. 2019{\natexlab{b}}.
\newblock Roberta: A robustly optimized bert pretraining approach.
\newblock \emph{arXiv preprint arXiv:1907.11692}.

\bibitem[{Mahabadi et~al.(2021{\natexlab{a}})Mahabadi, Henderson, and
  Ruder}]{mahabadi2021compacter}
Rabeeh~Karimi Mahabadi, James Henderson, and Sebastian Ruder.
  2021{\natexlab{a}}.
\newblock \href {https://openreview.net/forum?id=bqGK5PyI6-N} {Compacter:
  Efficient low-rank hypercomplex adapter layers}.
\newblock In \emph{NeurIPS}.

\bibitem[{Mahabadi et~al.(2021{\natexlab{b}})Mahabadi, Ruder, Dehghani, and
  Henderson}]{karimi-mahabadi-etal-2021-parameter_custom}
Rabeeh~Karimi Mahabadi, Sebastian Ruder, Mostafa Dehghani, and James Henderson.
  2021{\natexlab{b}}.
\newblock \href {https://aclanthology.org/2021.acl-long.47}
  {Parameter-efficient multi-task fine-tuning for transformers via shared
  hypernetworks}.
\newblock In \emph{ACL}.

\bibitem[{McCann et~al.(2018)McCann, Keskar, Xiong, and
  Socher}]{mccann2018natural}
Bryan McCann, Nitish~Shirish Keskar, Caiming Xiong, and Richard Socher. 2018.
\newblock \href {https://arxiv.org/abs/1806.08730} {The natural language
  decathlon: Multitask learning as question answering}.
\newblock \emph{arXiv preprint arXiv:1806.08730}.

\bibitem[{Min et~al.(2021)Min, Lewis, Zettlemoyer, and
  Hajishirzi}]{min2021metaicl}
Sewon Min, Mike Lewis, Luke Zettlemoyer, and Hannaneh Hajishirzi. 2021.
\newblock \href {https://arxiv.org/abs/2110.15943} {Meta{ICL}: Learning to
  learn in context}.
\newblock In \emph{NAACL}.

\bibitem[{Mishra et~al.(2022)Mishra, Khashabi, Baral, and
  Hajishirzi}]{mishra-etal-2022-cross}
Swaroop Mishra, Daniel Khashabi, Chitta Baral, and Hannaneh Hajishirzi. 2022.
\newblock \href {https://aclanthology.org/2022.acl-long.244} {Cross-task
  generalization via natural language crowdsourcing instructions}.
\newblock In \emph{ACL}.

\bibitem[{Mosbach et~al.(2021)Mosbach, Andriushchenko, and
  Klakow}]{mosbach2021on}
Marius Mosbach, Maksym Andriushchenko, and Dietrich Klakow. 2021.
\newblock \href {https://openreview.net/forum?id=nzpLWnVAyah} {On the stability
  of fine-tuning {BERT}: Misconceptions, explanations, and strong baselines}.
\newblock In \emph{ICLR}.

\bibitem[{Paszke et~al.(2019)Paszke, Gross, Massa, Lerer, Bradbury, Chanan,
  Killeen, Lin, Gimelshein, Antiga, Desmaison, Kopf, Yang, DeVito, Raison,
  Tejani, Chilamkurthy, Steiner, Fang, Bai, and
  Chintala}]{NEURIPS2019_bdbca288}
Adam Paszke, Sam Gross, Francisco Massa, Adam Lerer, James Bradbury, Gregory
  Chanan, Trevor Killeen, Zeming Lin, Natalia Gimelshein, Luca Antiga, Alban
  Desmaison, Andreas Kopf, Edward Yang, Zachary DeVito, Martin Raison, Alykhan
  Tejani, Sasank Chilamkurthy, Benoit Steiner, Lu~Fang, Junjie Bai, and Soumith
  Chintala. 2019.
\newblock \href
  {https://proceedings.neurips.cc/paper/2019/file/bdbca288fee7f92f2bfa9f7012727740-Paper.pdf}
  {Py{T}orch: An imperative style, high-performance deep learning library}.
\newblock In \emph{NeurIPS}.

\bibitem[{Pfeiffer et~al.(2021)Pfeiffer, Kamath, R{\"u}ckl{\'e}, Cho, and
  Gurevych}]{pfeiffer-etal-2021-adapterfusion}
Jonas Pfeiffer, Aishwarya Kamath, Andreas R{\"u}ckl{\'e}, Kyunghyun Cho, and
  Iryna Gurevych. 2021.
\newblock \href {https://aclanthology.org/2021.eacl-main.39}
  {{A}dapter{F}usion: Non-destructive task composition for transfer learning}.
\newblock In \emph{EACL}.

\bibitem[{Pilehvar and
  Camacho-Collados(2019)}]{pilehvar-camacho-collados-2019-wic}
Mohammad~Taher Pilehvar and Jose Camacho-Collados. 2019.
\newblock \href {https://aclanthology.org/N19-1128} {{W}i{C}: the
  word-in-context dataset for evaluating context-sensitive meaning
  representations}.
\newblock In \emph{NAACL}.

\bibitem[{Ponti et~al.(2022)Ponti, Sordoni, and Reddy}]{ponti2022combining}
Edoardo~M Ponti, Alessandro Sordoni, and Siva Reddy. 2022.
\newblock \href {https://arxiv.org/abs/2202.13914} {Combining modular skills in
  multitask learning}.
\newblock \emph{arXiv preprint arXiv:2202.13914}.

\bibitem[{Qin and Eisner(2021)}]{qin-eisner-2021-learning}
Guanghui Qin and Jason Eisner. 2021.
\newblock \href {https://aclanthology.org/2021.naacl-main.410} {Learning how to
  ask: Querying {LM}s with mixtures of soft prompts}.
\newblock In \emph{NAACL}.

\bibitem[{Radford et~al.(2021)Radford, Kim, Hallacy, Ramesh, Goh, Agarwal,
  Sastry, Askell, Mishkin, Clark et~al.}]{radford2021learning}
Alec Radford, Jong~Wook Kim, Chris Hallacy, Aditya Ramesh, Gabriel Goh,
  Sandhini Agarwal, Girish Sastry, Amanda Askell, Pamela Mishkin, Jack Clark,
  et~al. 2021.
\newblock \href {https://arxiv.org/abs/2103.00020} {Learning transferable
  visual models from natural language supervision}.
\newblock In \emph{ICML}.

\bibitem[{Raffel et~al.(2020)Raffel, Shazeer, Roberts, Lee, Narang, Matena,
  Zhou, Li, and Liu}]{raffel2019exploring}
Colin Raffel, Noam Shazeer, Adam Roberts, Katherine Lee, Sharan Narang, Michael
  Matena, Yanqi Zhou, Wei Li, and Peter~J. Liu. 2020.
\newblock \href {http://jmlr.org/papers/v21/20-074.html} {Exploring the limits
  of transfer learning with a unified text-to-text transformer}.
\newblock \emph{JMLR}.

\bibitem[{Rajpurkar et~al.(2016)Rajpurkar, Zhang, Lopyrev, and
  Liang}]{rajpurkar-etal-2016-squad}
Pranav Rajpurkar, Jian Zhang, Konstantin Lopyrev, and Percy Liang. 2016.
\newblock \href {https://aclanthology.org/D16-1264} {{SQ}u{AD}: 100,000+
  questions for machine comprehension of text}.
\newblock In \emph{EMNLP}.

\bibitem[{R{\"u}ckl{\'e} et~al.(2021)R{\"u}ckl{\'e}, Geigle, Glockner, Beck,
  Pfeiffer, Reimers, and Gurevych}]{ruckle-etal-2021-adapterdrop}
Andreas R{\"u}ckl{\'e}, Gregor Geigle, Max Glockner, Tilman Beck, Jonas
  Pfeiffer, Nils Reimers, and Iryna Gurevych. 2021.
\newblock \href {https://aclanthology.org/2021.emnlp-main.626} {{AdapterDrop}:
  {O}n the efficiency of adapters in transformers}.
\newblock In \emph{EMNLP}.

\bibitem[{Sakaguchi et~al.(2020)Sakaguchi, Le~Bras, Bhagavatula, and
  Choi}]{sakaguchi2020winogrande}
Keisuke Sakaguchi, Ronan Le~Bras, Chandra Bhagavatula, and Yejin Choi. 2020.
\newblock \href {https://arxiv.org/abs/1907.10641} {Wino{G}rande: An
  adversarial winograd schema challenge at scale}.
\newblock In \emph{AAAI}.

\bibitem[{Sanh et~al.(2022)Sanh, Webson, Raffel, Bach, Sutawika, Alyafeai,
  Chaffin, Stiegler, Raja, Dey, Bari, Xu, Thakker, Sharma, Szczechla, Kim,
  Chhablani, Nayak, Datta, Chang, Jiang, Wang, Manica, Shen, Yong, Pandey,
  Bawden, Wang, Neeraj, Rozen, Sharma, Santilli, Fevry, Fries, Teehan, Scao,
  Biderman, Gao, Wolf, and Rush}]{sanh2022multitask}
Victor Sanh, Albert Webson, Colin Raffel, Stephen Bach, Lintang Sutawika, Zaid
  Alyafeai, Antoine Chaffin, Arnaud Stiegler, Arun Raja, Manan Dey, M~Saiful
  Bari, Canwen Xu, Urmish Thakker, Shanya~Sharma Sharma, Eliza Szczechla,
  Taewoon Kim, Gunjan Chhablani, Nihal Nayak, Debajyoti Datta, Jonathan Chang,
  Mike Tian-Jian Jiang, Han Wang, Matteo Manica, Sheng Shen, Zheng~Xin Yong,
  Harshit Pandey, Rachel Bawden, Thomas Wang, Trishala Neeraj, Jos Rozen,
  Abheesht Sharma, Andrea Santilli, Thibault Fevry, Jason~Alan Fries, Ryan
  Teehan, Teven~Le Scao, Stella Biderman, Leo Gao, Thomas Wolf, and Alexander~M
  Rush. 2022.
\newblock \href {https://openreview.net/forum?id=9Vrb9D0WI4} {Multitask
  prompted training enables zero-shot task generalization}.
\newblock In \emph{ICLR}.

\bibitem[{Socher et~al.(2013)Socher, Perelygin, Wu, Chuang, Manning, Ng, and
  Potts}]{socher-etal-2013-recursive}
Richard Socher, Alex Perelygin, Jean Wu, Jason Chuang, Christopher~D. Manning,
  Andrew Ng, and Christopher Potts. 2013.
\newblock \href {https://aclanthology.org/D13-1170} {Recursive deep models for
  semantic compositionality over a sentiment treebank}.
\newblock In \emph{EMNLP}.

\bibitem[{Sung et~al.(2022)Sung, Cho, and Bansal}]{sung2022lst}
Yi-Lin Sung, Jaemin Cho, and Mohit Bansal. 2022.
\newblock Lst: Ladder side-tuning for parameter and memory efficient transfer
  learning.
\newblock \emph{arXiv preprint arXiv:2206.06522}.

\bibitem[{Talmor and Berant(2019)}]{talmor-berant-2019-multiqa}
Alon Talmor and Jonathan Berant. 2019.
\newblock \href {https://aclanthology.org/P19-1485} {{M}ulti{QA}: An empirical
  investigation of generalization and transfer in reading comprehension}.
\newblock In \emph{ACL}.

\bibitem[{Trischler et~al.(2017)Trischler, Wang, Yuan, Harris, Sordoni,
  Bachman, and Suleman}]{trischler-etal-2017-newsqa}
Adam Trischler, Tong Wang, Xingdi Yuan, Justin Harris, Alessandro Sordoni,
  Philip Bachman, and Kaheer Suleman. 2017.
\newblock \href {https://aclanthology.org/W17-2623} {{N}ews{QA}: A machine
  comprehension dataset}.
\newblock In \emph{Proceedings of the 2nd Workshop on Representation Learning
  for {NLP}}.

\bibitem[{Vu et~al.(2022)Vu, Lester, Constant, Al-Rfou{'}, and
  Cer}]{vu2021spot}
Tu~Vu, Brian Lester, Noah Constant, Rami Al-Rfou{'}, and Daniel Cer. 2022.
\newblock \href {https://aclanthology.org/2022.acl-long.346} {{SP}o{T}: Better
  frozen model adaptation through soft prompt transfer}.
\newblock In \emph{ACL}.

\bibitem[{Vu et~al.(2020)Vu, Wang, Munkhdalai, Sordoni, Trischler,
  Mattarella-Micke, Maji, and Iyyer}]{vu-etal-2020-exploring}
Tu~Vu, Tong Wang, Tsendsuren Munkhdalai, Alessandro Sordoni, Adam Trischler,
  Andrew Mattarella-Micke, Subhransu Maji, and Mohit Iyyer. 2020.
\newblock \href {https://aclanthology.org/2020.emnlp-main.635} {Exploring and
  predicting transferability across {NLP} tasks}.
\newblock In \emph{EMNLP}.

\bibitem[{Wang et~al.(2019{\natexlab{a}})Wang, Pruksachatkun, Nangia, Singh,
  Michael, Hill, Levy, and Bowman}]{wang2019superglue}
Alex Wang, Yada Pruksachatkun, Nikita Nangia, Amanpreet Singh, Julian Michael,
  Felix Hill, Omer Levy, and Samuel Bowman. 2019{\natexlab{a}}.
\newblock \href
  {https://proceedings.neurips.cc/paper/2019/file/4496bf24afe7fab6f046bf4923da8de6-Paper.pdf}
  {Superglue: A stickier benchmark for general-purpose language understanding
  systems}.
\newblock In \emph{NeurIPS}.

\bibitem[{Wang et~al.(2019{\natexlab{b}})Wang, Singh, Michael, Hill, Levy, and
  Bowman}]{wang-etal-2018-glue}
Alex Wang, Amanpreet Singh, Julian Michael, Felix Hill, Omer Levy, and
  Samuel~R. Bowman. 2019{\natexlab{b}}.
\newblock \href {https://openreview.net/forum?id=rJ4km2R5t7} {{GLUE}: A
  multi-task benchmark and analysis platform for natural language
  understanding}.
\newblock In \emph{ICLR}.

\bibitem[{Wang et~al.(2022{\natexlab{a}})Wang, Mishra, Alipoormolabashi, Kordi,
  Mirzaei, Arunkumar, Ashok, Dhanasekaran, Naik, Stap
  et~al.}]{wang2022benchmarking}
Yizhong Wang, Swaroop Mishra, Pegah Alipoormolabashi, Yeganeh Kordi, Amirreza
  Mirzaei, Anjana Arunkumar, Arjun Ashok, Arut~Selvan Dhanasekaran, Atharva
  Naik, David Stap, et~al. 2022{\natexlab{a}}.
\newblock \href {https://arxiv.org/abs/2204.07705} {Benchmarking generalization
  via in-context instructions on 1,600+ language tasks}.
\newblock \emph{arXiv preprint arXiv:2204.07705}.

\bibitem[{Wang and Chen(2020)}]{wang-chen-2020-position}
Yu-An Wang and Yun-Nung Chen. 2020.
\newblock \href {https://aclanthology.org/2020.emnlp-main.555} {What do
  position embeddings learn? an empirical study of pre-trained language model
  positional encoding}.
\newblock In \emph{EMNLP}.

\bibitem[{Wang et~al.(2022{\natexlab{b}})Wang, Zhang, Lee, Zhang, Sun, Ren, Su,
  Perot, Dy, and Pfister}]{wang2022learning}
Zifeng Wang, Zizhao Zhang, Chen-Yu Lee, Han Zhang, Ruoxi Sun, Xiaoqi Ren,
  Guolong Su, Vincent Perot, Jennifer Dy, and Tomas Pfister.
  2022{\natexlab{b}}.
\newblock Learning to prompt for continual learning.
\newblock In \emph{CVPR}.

\bibitem[{Warstadt et~al.(2019)Warstadt, Singh, and
  Bowman}]{warstadt2019neural}
Alex Warstadt, Amanpreet Singh, and Samuel~R. Bowman. 2019.
\newblock \href {https://aclanthology.org/Q19-1040} {Neural network
  acceptability judgments}.
\newblock \emph{TACL}.

\bibitem[{Wei et~al.(2022)Wei, Bosma, Zhao, Guu, Yu, Lester, Du, Dai, and
  Le}]{wei2022finetuned}
Jason Wei, Maarten Bosma, Vincent Zhao, Kelvin Guu, Adams~Wei Yu, Brian Lester,
  Nan Du, Andrew~M. Dai, and Quoc~V Le. 2022.
\newblock \href {https://openreview.net/forum?id=gEZrGCozdqR} {Finetuned
  language models are zero-shot learners}.
\newblock In \emph{ICLR}.

\bibitem[{Williams et~al.(2018)Williams, Nangia, and
  Bowman}]{williams-etal-2018-broad}
Adina Williams, Nikita Nangia, and Samuel Bowman. 2018.
\newblock \href {https://aclanthology.org/N18-1101} {A broad-coverage challenge
  corpus for sentence understanding through inference}.
\newblock In \emph{NAACL}.

\bibitem[{Wolf et~al.(2020)Wolf, Debut, Sanh, Chaumond, Delangue, Moi, Cistac,
  Rault, Louf, Funtowicz, Davison, Shleifer, von Platen, Ma, Jernite, Plu, Xu,
  Le~Scao, Gugger, Drame, Lhoest, and Rush}]{wolf-etal-2020-transformers}
Thomas Wolf, Lysandre Debut, Victor Sanh, Julien Chaumond, Clement Delangue,
  Anthony Moi, Pierric Cistac, Tim Rault, Remi Louf, Morgan Funtowicz, Joe
  Davison, Sam Shleifer, Patrick von Platen, Clara Ma, Yacine Jernite, Julien
  Plu, Canwen Xu, Teven Le~Scao, Sylvain Gugger, Mariama Drame, Quentin Lhoest,
  and Alexander Rush. 2020.
\newblock \href {https://aclanthology.org/2020.emnlp-demos.6} {Transformers:
  State-of-the-art natural language processing}.
\newblock In \emph{EMNLP: System Demonstrations}.

\bibitem[{Wu et~al.(2022)Wu, Wang, Gu, Hou, Dong, Vydiswaran, and
  Ma}]{wu2022idpg}
Zhuofeng Wu, Sinong Wang, Jiatao Gu, Rui Hou, Yuxiao Dong, VG~Vydiswaran, and
  Hao Ma. 2022.
\newblock \href {https://arxiv.org/abs/2204.04497?context=cs.LG} {{IDPG}: An
  instance-dependent prompt generation method}.
\newblock \emph{arXiv preprint arXiv:2204.04497}.

\bibitem[{Yang et~al.(2018)Yang, Qi, Zhang, Bengio, Cohen, Salakhutdinov, and
  Manning}]{yang-etal-2018-hotpotqa}
Zhilin Yang, Peng Qi, Saizheng Zhang, Yoshua Bengio, William Cohen, Ruslan
  Salakhutdinov, and Christopher~D. Manning. 2018.
\newblock \href {https://aclanthology.org/D18-1259} {{H}otpot{QA}: A dataset
  for diverse, explainable multi-hop question answering}.
\newblock In \emph{EMNLP}.

\bibitem[{Zhang et~al.(2018)Zhang, Liu, Liu, Gao, Duh, and
  Van~Durme}]{zhang2018record}
Sheng Zhang, Xiaodong Liu, Jingjing Liu, Jianfeng Gao, Kevin Duh, and Benjamin
  Van~Durme. 2018.
\newblock \href {https://arxiv.org/abs/1810.12885} {Record: Bridging the gap
  between human and machine commonsense reading comprehension}.
\newblock \emph{arXiv preprint arXiv:1810.12885}.

\bibitem[{Zhang et~al.(2020)Zhang, Deng, Zhang, and Wu}]{zhang2020survey}
Wen Zhang, Lingfei Deng, Lei Zhang, and Dongrui Wu. 2020.
\newblock \href {https://arxiv.org/abs/2009.00909} {A survey on negative
  transfer}.
\newblock \emph{arXiv preprint arXiv:2009.00909}.

\bibitem[{Zhang et~al.(2015)Zhang, Zhao, and LeCun}]{zhang2015character}
Xiang Zhang, Junbo Zhao, and Yann LeCun. 2015.
\newblock \href
  {https://proceedings.neurips.cc/paper/2015/file/250cf8b51c773f3f8dc8b4be867a9a02-Paper.pdf}
  {Character-level convolutional networks for text classification}.
\newblock In \emph{NeurIPS}.

\bibitem[{Zhang et~al.(2019)Zhang, Baldridge, and He}]{zhang-etal-2019-paws}
Yuan Zhang, Jason Baldridge, and Luheng He. 2019.
\newblock \href {https://aclanthology.org/N19-1131} {{PAWS}: Paraphrase
  adversaries from word scrambling}.
\newblock In \emph{NAACL}.

\end{thebibliography}
\bibliographystyle{acl_natbib}

\clearpage
\appendix

\section*{Appendix}

\label{sec:appendix}
\section{More Method Details}

\subsection{Improving Multi-task Training}

{
Learning effective interpolations of prompts is challenging, as input embeddings themselves do not necessarily correspond to meaningful prompt tokens, and we do not have any supervisions for the ground truth task mapping. 
}
We explore several approaches to improve the training with good inductive bias so that $\mathcal{G}$ learns a good prompt composition for efficient knowledge transfer.

\paragraph{Learning attention prior.} We pre-train the attention module on source tasks and then use the learned projection layers and the layer norm to initialize the attention module on the target task(s). 
This learned prior can be also directly used for tasks that lack training data. 

To learn the attention prior for $\mathcal{G}$, we run the same training process as in the target task training on the source tasks. In particular, we initialize another task-specific prompt for each source task, and trains both those task-specific prompts as well as the shared attention weights of $\mathcal{G}$ on the combinations of the source tasks as in \Sref{sec:attention}.

\paragraph{Two-speed learning rate.}
\citet{ponti2022combining} shows that setting different learning rates for the composition module and the task-specific model parameters helps to provide useful inductive bias to encourage the model to learn the best skill composition. 
We also introduce this two-speed learning rate approach for \ours.

\subsection{Overview of Training}
Algorithm Table~\ref{tab:algo_box} presents the overview of the training algorithm. 

\subsection{Parameter Efficiency of {\ours}}
\begin{figure}[t!]
\centering
\includegraphics[width=5cm]{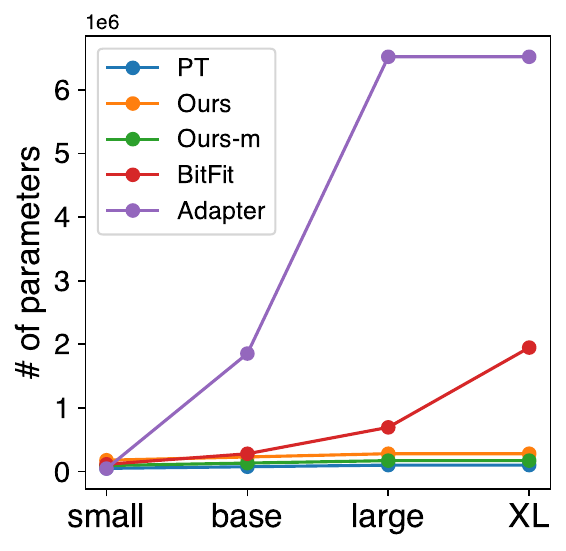}\caption{The number of the parameters to be updated with Adapter, BitFit, Fine-tuning, Prompt Tuning and ours using different backbone LMs. Ours and Ours-m denote {\ours} and {\ours}-m, respectively. 
} \label{fig:params}
\end{figure}

\Fref{fig:params} shows the number of the parameters to be updated for Prompt tuning, {\ours}, Adapter, and BitFit when we increase the size of the backbone LMs. 
As we can see, other parameter-efficient transfer approaches observe quick increases of the trainable parameters, while {\ours} shows small increases. In addition, {\ours} keeps the original LM frozen and does not modify the LM structures unlike those approaches. 

\subsection{Alternative Attention Design}
{\ours} computes the same attention scores over $m$ prompt tokens. 
Alternatively, we compute attention scores for {\it each} prompt token further flexibility and expressiveness. Here, instead of computing similarities between the summary representation of $\mathbf{H}_{out}$ and prompt $\hat{\mathbf{P}_j}$, we compute similarities between $\mathbf{H}_{out}$ and each $l$th prompt token as follows:
\begin{equation}\label{eq:soft_max_token}
    a_{lj} = \frac{e^{{\mathbf{p}_{lj}} \mathbf{H}_{out}}}{\sum_{k=1}^{t+1} e^{{\mathbf{p}_{lk}} \mathbf{H}_{out}}}.
\end{equation}
For prompt token-level attention, in the second term on the right, each $l$th prompt token in the summary representation is calculated as the weighted summary of the $l$th prompt tokens. 

Empirically the token-level attentions gives similar performance to the original attention in \Eref{eq:soft_max}, while in some tasks it gives notable performance improvements. Due to the additional computational overhead, we use the max-pooling based unified attention (\Eref{eq:soft_max}) as our default attention mechanism. 
Interestingly, we find that the attention distributions are significantly different among prompt tokens in different locations (e.g., giving significantly higher attentions to the target task prompt in the later tokens), potentially because of the position biases of pretrained models~\cite{wang-chen-2020-position}. 
This is beyond the scope of this work, but is certainly of interest for future work. 

\begin{table*}[t!]
\begin{tcolorbox}
{\bf Source Prompt Training }\\ 
For $j$th source tasks in $t$ source tasks, train a source prompt  $\mathbf{P}_j$ by maximizing $p(\boldsymbol{y} \mid [\mathbf{P}_j, \mathbf{X}]$ individually {\bf (\Sref{sec:source_target})~[\Eref{eq:training}] }
\tcblower
{\bf Target Prompt Training}  \\
{\bf Initialization:} initialize a new prompt $\mathbf{P}_{\it target}$ and attention module $\mathcal{G}$ \\ 
For each instance $(\boldsymbol{x}, \boldsymbol{y})$, after passing $\boldsymbol{x}$ to the embedding layer to get input embeddings $\mathbf{X}$, \\
{\bf Step 1:} Compute instance-wise prompt $\mathbf{P}_{\it instance}$ for $\mathbf{X}$ {\bf (\Sref{sec:attention})}
\begin{enumerate}
\itemsep0em 
    \item calculate attentions between $\mathbf{X}$ and a set of prompts $[\mathbf{P}_1, \ldots, \mathbf{P}_t, \mathbf{P}_{\it target}]$ using $\mathcal{G}$  {\bf [\Eref{eq:soft_max}]}
    \item interpolate $\mathbf{P}_1, \ldots \mathbf{P}_t$ and $\mathbf{P}_{\it target}$ using attention scores {\bf [\Eref{eq:final_soft}] }
\end{enumerate}
{\bf Step 2:} Prepend  $\mathbf{P}_{\it instance}$ to $\mathbf{X}$ and feed the final input to frozen LM $\theta$ \\
{\bf Step 3:} Maximize $p(\boldsymbol{y} \mid [\mathbf{P}_{\it instance}, \mathbf{X}])$ and backpropagate to $\mathbf{P}_{\it target}$ and $\mathcal{G}$ via $\mathbf{P}_{\it instance}$~{\bf [\Eref{eq:training}]}
\end{tcolorbox}
\caption{\label{tab:algo_box} Training process of \ours. 
}
\end{table*}

\section{Task and Dataset Details}
\label{sec:dataset_appendix}
We show the list of the datasets, tasks and domains for source tasks in \Tref{tab:task_overview_source} and for target tasks in \Tref{tab:task_overview_target}, respectively. 
In summary, both source and target datasets cover diverse tasks, domains and output formats (i.e., span extraction, multiple-choice, classification).  

\section{Experimental Details}

\subsection{Implementation Details}
We use PyTorch\footnote{\url{https://pytorch.org/}}~\cite{NEURIPS2019_bdbca288} and huggingface \texttt{transformers}\footnote{\url{https://github.com/huggingface/transformers}}~\cite{wolf-etal-2020-transformers} to implement our models.
For Adapter, BitFit, prompt tuning and BitFit baselines, we use the implementations by \citet{mahabadi2021compacter}.\footnote{\url{https://github.com/rabeehk/compacter}}
We use huggingface \texttt{datasets}\footnote{\url{https://github.com/huggingface/datasets}} library to use the data for the experiments except for MRQA 2019 shared task. For MRQA 2019 shared task, we download the original training and development data from the official repository.\footnote{\url{https://github.com/mrqa/MRQA-Shared-Task-2019}} 

\subsection{Source Prompt Training Details}
\label{sec:dataset_details}
We fine-tune the source prompts on six large-scale datasets for 5 epochs. We use the checkpoints with the best development score as our source prompts. 
Each source prompt is initialized by randomly sampled tokens as in \citet{lester-etal-2021-power}. We found that although this random vocabulary based initialization is often unstable even in large-scale datasets, on the six source tasks, this approach gives reasonable performance, even with T5-small.  

\subsection{Attention Module Pretraining Details}
As the six source tasks have significantly different length of input context (e.g., the input context of MNLI, SST-2, QQP or QNLI is on average less than 200 tokens while SQuAD or ReCoRD have the context longer than 512 tokens), we split the source tasks into the two groups: (1) MNLI, SST-2, QQP and QNLI; (2) SQuAD and ReCoRD. We use the resulting pretrained weights from group (2) for MRQA 2019, while for other experiments, we use the weights from (1).

\begin{table*}[t!]
\small
    \centering
    \begin{tabular}{l| llll}
\toprule
Dataset Name & Category & Task & Domain & Metric  \\\midrule
1. MNLI & GLUE & natural language inference (NLI) & various & accuracy   \\
2. SST-2 & GLUE & sentiment analysis & Movie Reviews & accuracy  \\
3. QQP & GLUE & paraphrase detection & social QA questions (Quora) & accuracy \& F1 \\
4. QNLI & GLUE QA & NLI & Wikipedia & accuracy \\
5. SQuAD & MRQA 2019 & extractive QA & Wikipedia & F1 \& EM \\
6. ReCoRD & SuperGLUE & cloze-style QA & news (CNN, Daily Mail) & F1 \& EM \\
 \bottomrule
 \end{tabular}
    \caption{The details of the 6 source tasks. MNLI, SST-2, QQP and QNLI are also used as target tasks in GLUE experiments.
    }
    \label{tab:task_overview_source}
\end{table*}
\begin{table*}[t!]
\small
    \centering
    \begin{tabular}{l| llll}
\toprule
Dataset Name & Category & Task & Domain & Metric \\\midrule
1. CoLA & GLUE & acceptability & various & Matthews corr.  \\
2. STS-B &  GLUE & sentence similarity &  various & \underline{Pearson}\&Spearman corr. \\
3. MRPC & GLUE & paraphrase detection  & news & \underline{accuracy} \& {F1}  \\
4. RTE & GLUE & NLI & News, Wikipedia & accuracy \\
5. MultiRC & SuperGLUE & QA & various & \underline{F1} \& EM \\
6. BoolQ & SuperGLUE & boolean QA &  Wikipedia & accuracy   \\
7. WiC & SuperGLUE & word sense disambiguation  & lexical databases & accuracy \\
8. WSC & SuperGLUE & coreference / commonsense & fiction books & accuracy\\
9. CB & SuperGLUE & NLI & various & accuracy \\
10. NQ & MRQA 2019 & extractive QA & Wikipedia & \underline{F1} \& EM   \\
11. HotpotQA & MRQA 2019 & extractive QA & Wikipedia & \underline{F1} \& EM\\
12. SearchQA & MRQA 2019 & extractive QA & Search snippets & \underline{F1} \& EM\\
13. NewsQA & MRQA 2019 & extractive QA & News article & \underline{F1} \& EM \\
14. WinoGrande & Others & coreference / commonsense & WikiHow & accuracy  \\
15. Yelp & Others & sentiment analysis & Yelp reviews & accuracy \\
16. SciTail & Others & NLI & science exams & accuracy \\
17. PAWS-Wiki & Others & paraphrase detection & Wikipedia & accuracy \\
 \bottomrule
 \end{tabular}
    \caption{The details of the 17 target tasks except for 4 GLUE datasets, which are also used for evaluation. ``NQ'' denotes Natural Questions and lexical databases for WiC include WordNet, VerbNet, Wiktionary. For the datasets where two metrics are originally used, we use the underlined metric as our primary metric. 
    }
    \label{tab:task_overview_target}
\end{table*}

\subsection{General hyperparameters}
We set the maximum token length to be 512 for MRQA datasets, 348 for MultiRC and 256 for all of other datasets.
All of the experiments are conducted with a single GPU with 24 GB memory. On all of the datasets, training were completed within 24 hours.
Per GPU batch size is 32, and for MRQA, we set the per GPU batchsize to be 16 and set the gradient accumulation step to 2 due to the out of memory error. 

\subsection{Hyperparameters for \ours}
We use $T= d \times {\rm exp}(1)$, where $d$ is the LM dimension size, to control the soft max temperature in \Sref{sec:attention}.
The prompt length $m$ is 100 and the prompt tuning learning rate is 0.3 and optimize the objective function using Adam~\cite{kingma2014adam}. 
We set weight decay to be $1 \times 10^{-5}$. 
For the projection layers, we use $r=100$. 
For the attention module $\mathcal{G}$, we found that the best learning rate varies across datasets and tune it on the development sets. In particular, we use the learning rate of 0.1 for SuperGLUE, and Yelp, WinoGrande, SciTail and PAWS multi-task experiments, and 0.3 for the other experiments.

\subsection{Hyperparameters for Baselines}
For all of the baselines, we set the warmup steps to be 500, use Adam for optimization with a linear learning rate scheduler. 
\paragraph{Prompt Tuning.}
As in \ours, we use the prompt length of $m=100$ and use the learning rate of 0.3 for prompt tuning and set weight decay to be $1 \times 10^{-5}$. 

\paragraph{SPoT.}
We explore two approaches to initalize the target task prompt as in \citet{vu2021spot}: {\bf SPoT-{\it generic} (SPoT-g)} and {\bf SPoT-{\it targeted}} ({\bf SPoT-t}). 
SPoT-g first pre-trains source prompts on eight GLUE tasks and then uses the source prompts to initialize target task prompts, while SPoT-t uses prompt similarities to find top-$k$ similar tasks and then initializes target task prompts using the top $k$ prompts. As we only use 6 source tasks in this work, we use top 1 similar prompt as the transfer source of SPoT-t. 
We use the same hyperparameters as in prompt tuning. To select the source task for SPoT-t, we run prompt tuning on all of the source and target tasks for 5 epochs for medium and large-scale datasets and 20 epochs for smaller scale datasets and then compute the cosine similarity between a target prompt and the set of the source prompts. 
Regarding the SPoT-g training, we train a single source prompt on the combination of the GLUE source tasks following \citet{vu2021spot}. 
We found that SPoT-g baseline is not strong on MRQA or Others (i.e., Yelp, Scitail, WinoGrande and PAWS-Wiki), while it gives small performance improvements on GLUE from SPoT-t in some tasks. Therefore, we use SPoT-t in our main experiments. 

\paragraph{Adapter.}
We use the default hyperparameters by \citet{mahabadi2021compacter} for the Adapter baseline. We use GELU~\cite{hendrycks2016gaussian} for non-linear layers, set the reduction factor to be 32 and the learning rate to be $3 \times 10^{-4}$.

\paragraph{BitFit.}
We use the learning rate of $3 \times 10^{-4}$.

\paragraph{Fine-tuning.}
We use the learning rate of $3 \times 10^{-4}$. Other hyperparameters are the same as the huggingface transformers T5 models.

\subsection{Multi-task Training Details}
\label{sec:multi_task_details}
The 17 datasets have significantly different length of input context, and training on the combinations of all of the datasets can make training inefficient. We conduct multi-tasking of 4 datasets (SuperGLUE, MRQA 2019, and others), while on GLUE, we train \ours-m on 8 GLUE tasks. 
We keep MultiRC training separated from other SuperGLUE tasks, as MultiRC has significantly longer context than other SuperGLUE datasets. 
We set the maximum length of the input to be 256, 256, 512, 256 for GLUE, SuperGLUE, MRQA 2019, and others task set, respectively. We set the maximum length of input to be 348 for MultiRC.

\subsection{Few-shot Adaptation Experiments Details}
Following~\citet{karimi-mahabadi-etal-2021-parameter_custom}, we run few-shot adaptation experiments for three times and takes the mean of the performance. We cite the performance of the fine-tuning, Adapter and HyperFormer from \citet{karimi-mahabadi-etal-2021-parameter_custom}, and train a single prompt tuning model on 8 GLUE tasks and then transfer it to few-shot tasks. For {\ours}, we load the attention weights trained on 8 GLUE tasks.

\subsection{Scaling Experiments Details}
\begin{table}[t!]
\small
    \centering
    \begin{tabular}{l| ccc |ccc }
\toprule 
& \multicolumn{3}{c}{Adapter} & \multicolumn{3}{c}{Fine-tuning} \\
datasets & Bool & MRC & WiC & Bool & MRC & WiC \\\midrule
T5-small & 100 & 100 & 100 & 100 & 100 & 100  \\
T5-base & 64 & 64 & 100  & 32 & 32& 100 \\
T5-large & 32 & 20 & 32   & 32 & 32 &  32 \\
T5-3B & 4 & 4 & 8 & -- & -- & --\\
 \bottomrule
 \end{tabular}
    \caption{The number of the batch sizes for fine-tuned models and adapter for the scalability experiments.  
    }
    \label{tab:per_device_batchsize}
\end{table}
During this experiment, we use only a single GPU with 24 GB GPU memory, as in our main experiments, to simulate a common resource environment.
We found that under this computational constraint, we could not fine-tune the T5-3B model due to the out of memory error, even with a batch size of 1. 
Adapter, prompt tuning and \ours~can be trained on a single GPU even with the T5-3B model. 
We provide the experimental details for the LM scaling experiments in \Sref{sec:increase_base_size}. 
For \ours~and prompt tuning, we use the same single GPU with 24 GB GPU memory as the main experiments. For Adapter and fine-tuning, we use a single GPU with 48 GB GPU memory but restrict GPU memory usage at 24 GB for a fair comparison. 
For the scalability experiments, we set the maximum token length to 216 across all datasets.  

\paragraph{Per-device batch size for \ours~and prompt tuning.}
For T5 small and base, we set per-GPU batch size to be 100 and 32, while for T5-large and T5-XL (3B), we use the batch size of 16 and 2, respectively.

\paragraph{Per-device batch size for Adapter.}
For Adapter experiments, we flexibly adjust the per-device batch size for each dataset to avoid out of the memory issues. 
The number of the per-device batch size is shown in \Tref{tab:per_device_batchsize}. 

\paragraph{Per-device batch size for fine-tuning.}
Similarly in Adapter, we adjust the per-device batch size for the fine-tuned models. The number of the per-device batch size is shown in \Tref{tab:per_device_batchsize}. 
For fine-tuned models, we found that we cannot avoid the out of memory issue even with the batch size of 1, so we report the results with T5 small, base and large.

\paragraph{Performance instability of fine-tuning with T5-large.}
We found that fine-tuning with T5-large is occasionally unstable and fails to learn a target task, and is sensitive to the batch size or learning rate. 
For instance, using different batch size results in 65\% BoolQ accuracy. 
For those cases, we explored several learning rates and batch sizes and report the best performance. Several prior work report the instability of fine-tuning large-scale LMs~\cite{mosbach2021on,dodge2020fine}.

\subsection{Memory Footprints}
\label{sec:memory_footprint}
{
Despite its parameter-efficiency, prompt tuning based approaches increase the sequence length by prepending continuous emeddings in front of the original input sequence~\cite{lester-etal-2021-power,mahabadi2021compacter}.
We evaluate the memory footprint of full fine-tuning, Adapter, BitFit, prompt tuning and \ours. We use  T5-base as a default base LM and set the per-gpu batch size to 32. 
We also compare the memory footprint using T5-3B with batch size of 2. We set the length of the prompt to 100.} 

{
As shown in Table~\ref{tab:memory_footprint}, \ours~increase the memory footprint from other methods, due to the increase of the input length, multiple pre-loaded source prompts and attention calculations.  
On the other hand, \ours~shows moderate memory footprint increase when the backbone LM size gets larger (13.7 GB to 16.1 GB) while Adapter and BitFit show about three times more memory footprints than T5-base. 
This demonstrates that \ours~is more parameter-efficient and can be more memory-efficient when the backbone LMs get even larger (e.g., 11 billions).
Moreover, \citet{lester-etal-2021-power} show that the input prompt length can be significantly reduced when the backbone LMs get larger, which further improve the memory efficiency of prompt tuning-based methods. 
}
\begin{table}[t!]
\small
    \centering
    \begin{tabular}{l|cc}
\toprule 
 & \multicolumn{2}{c}{memory footprint } \\ 
 &  (base) &  (XL)    \\
\midrule
Fine-tuning & 9.0 GB & -- \\
Adapter & 5.9 GB & 14.5 GB \\
BitFit & 5.6 GB  & 14.2 GB \\
Prompt Tuning & 8.5 GB & 15.9 GB \\
\ours~(single) & 13.7 GB & 16.1 GB \\
 \bottomrule
 \end{tabular}
    \caption{The maximum memory footprint during training of fine-tuning, Adapter, BitFit, prompt tuning and \ours (single task).    
    }
    \label{tab:memory_footprint}
\end{table}

\end{document}